\documentclass[acmtog]{acmart}

\renewcommand\footnotetextcopyrightpermission[1]{}

\usepackage{booktabs} % For formal tables
\usepackage{xspace} 

\usepackage[ruled]{algorithm2e} % For algorithms

\SetAlFnt{\small}
\SetAlCapFnt{\small}
\SetAlCapNameFnt{\small}
\SetAlCapHSkip{0pt}
\usepackage[export]{adjustbox}
\usepackage{xspace}
\usepackage{array}
\newcolumntype{P}[1]{>{\centering\arraybackslash}p{#1}}
\usepackage{tikz}
\usepackage{amsmath}
\usepackage{newunicodechar}
\newunicodechar{↔}{\ensuremath{\leftrightarrow}}
\usetikzlibrary{arrows.meta, positioning, fit, shapes.geometric}
\usepackage{multirow}
\usepackage{graphicx}
\usepackage{standalone}
\usepackage{enumitem}
\usepackage{framed}

\begin{document}
% Title portion
\newcommand{\LightDome}{LED Sphere\xspace}
\newcommand{\shortname}{BodyReLux\xspace}
\title{\shortname: Temporally Consistent Full-Body Video Relighting}

\author[Li Ma]{Li Ma}
\orcid{0000-0002-6992-0089}
\affiliation{
  \institution{Eyeline Labs}
  \city{Los Angeles}
  \country{United States of America}
}
\email{li.ma@scanlinevfx.com}

\author[Mingming He]{Mingming He}
\orcid{0000-0002-9982-7934}
\affiliation{
  \institution{Eyeline Labs}
  \city{Los Angeles}
  \country{United States of America}
}
\email{mingming.he@scanlinevfx.com}

\author[Xueming Yu]{Xueming Yu}
\orcid{0009-0009-8189-6024}
\affiliation{
  \institution{Eyeline Labs}
  \city{Los Angeles}
  \country{United States of America}
}
\email{xueming.yu@scanlinevfx.com}

\author[David M. George]{David M. George}
\orcid{0009-0001-1570-3708}
\affiliation{
  \institution{Eyeline Labs}
  \city{Los Angeles}
  \country{United States of America}
}
\email{david.george@scanlinevfx.com}

\author[Ahmet Levent Ta\c{s}el]{Ahmet Levent Taşel}
\orcid{0009-0002-7150-0160}
\affiliation{
  \institution{Eyeline Labs}
  \city{Vancouver}
  \country{Canada}
}
\email{ahmet.tasel@scanlinevfx.com}

\author[Paul Debevec]{Paul Debevec}
\orcid{0000-0001-7381-2323}
\affiliation{
  \institution{Eyeline Labs and Netflix}
  \city{Los Angeles}
  \country{United States of America}
}
\email{debevec@gmail}

\author[Julien Philip]{Julien Philip}
\orcid{0000-0003-3125-1614}
\affiliation{
  \institution{Eyeline Labs}
  \city{London}
  \country{United Kingdom}
}
\email{julien.philip@scanlinevfx.com}

\newcommand{\TODO}[1]{{\bf\color{red}[TODO: #1]}}

\newcommand{\camready}[1]{#1}

\newcommand{\JULIENTEXT}[1]{{\bf\color{purple}#1}}
\newcommand{\JULIEN}[1]{{\bf\color{purple}[JULIEN: #1]}}
\newcommand{\LITEXT}[1]{{\color{cyan}#1}}
\newcommand{\li}[1]{{#1}}
\newcommand{\lirm}[1]{{\color{gray}#1}}
\newcommand{\AHMETTEXT}[1]{{\color{brown}#1}}
\newcommand{\AHMET}[1]{{\bf\color{brown}[AHMET: #1]}}
\newcommand{\WENQITEXT}[1]{{\color{purple}#1}}
\newcommand{\WENQI}[1]{{\bf\color{purple}[WENQI: #1]}}
\newcommand{\MINGMINGTEXT}[1]{{\bf\color{orange}#1}}
\newcommand{\mingming}[1]{{\bf\color{orange}[MINGMING: #1]}}
\newcommand{\PAULTEXT}[1]{{\color{teal}#1}}
\newcommand{\PAUL}[1]{{\bf\color{teal}[PAUL: #1]}}
\newcommand{\NINGTEXT}[1]{{\color{red}#1}}
\newcommand{\NING}[1]{{\bf\color{red}[NING: #1]}}

\begin{abstract}
Being able to relight human performance is a fundamental task for post production and content creation. 
We present \shortname, a subject-specific video diffusion-based framework for relighting full-body human performances in a temporally consistent way. 
% Our relighting model is trained using a novel sequence of paired lighting condition captured in a large scale LED sphere.  
% We have each subject perform a few-minute sequence of movements while recording with a small set of standard cinema cameras running at 120 frames per second.  As they perform, we interleave two gradually transforming spherical environment lighting patterns, sequences A and B, each displaying its next gradually changing pattern 60 times per second,  The lighting sequence exceeds the human visual system's 60Hz flicker fusion frequency and is comfortable to perform within for the capture duration; and the subject does not need to hold any poses.  After temporally aligning sequences A and B, we have a plethora of paired video clips of the subject performing the same aligned motions but recorded under two different lighting conditions (A and B). 
% We then train a diffusion-based video relighting model using a novel lighting conditioning method that treats each light source as a token. Together with a carefully designed data augmentation pipeline, we achieve photorealistic, robust, and temporally consistent video relighting of subject-specific human performances.
% \TODO{reduce the details of bi-pack.}
\li{Our model is trained on a hybrid dataset of pixel-aligned video relighting pairs, covering a diverse combination of lighting conditions, performances and viewpoints. 
To acquire such dataset, we combine traditional static One-Light-at-a-Time (OLAT) capture and a novel dynamic performance capture in which two smoothly varying lighting sequences are rapidly interleaved. 
Because the lighting operates above the human flicker-fusion threshold, the interleaving does not appears to strobe.
We train our video relighting model from a pretrained text-to-video model to fully leverage the generative priors for producing high quality videos.
To achieve accurate lighting control, we introduce a new lighting conditioning method that represents each light source as a token. We further condition on sequences of lighting using masked attention to support dynamic lighting control.
Together with a carefully designed data augmentation pipeline, we achieve photorealistic, robust, and temporally consistent video relighting of subject-specific human performances.
}

\end{abstract}

%    <concept>
%        <concept_id>10010147.10010371.10010372.10010373</concept_id>
%        <concept_desc>Computing methodologies~Rasterization</concept_desc>
%        <concept_significance>500</concept_significance>
%        </concept>
%    <concept>
%        <concept_id>10010147.10010371.10010382.10010385</concept_id>
%        <concept_desc>Computing methodologies~Image-based rendering</concept_desc>
%        <concept_significance>500</concept_significance>
%        </concept>
%    <concept>
%        <concept_id>10010147.10010371.10010382.10010383</concept_id>
%        <concept_desc>Computing methodologies~Image processing</concept_desc>
%        <concept_significance>500</concept_significance>
%        </concept>
%    <concept>
%        <concept_id>10010147.10010257.10010293</concept_id>
%        <concept_desc>Computing methodologies~Machine learning approaches</concept_desc>
%        <concept_significance>500</concept_significance>
%        </concept>
%  </ccs2012>

% \ccsdesc[500]{Computing methodologies~Rasterization}
% \ccsdesc[500]{Computing methodologies~Image-based rendering}
% \ccsdesc[500]{Computing methodologies~Image processing}
% \ccsdesc[500]{Computing methodologies~Machine learning approaches}



\begin{teaserfigure}
\centering
\includegraphics[width=\textwidth]{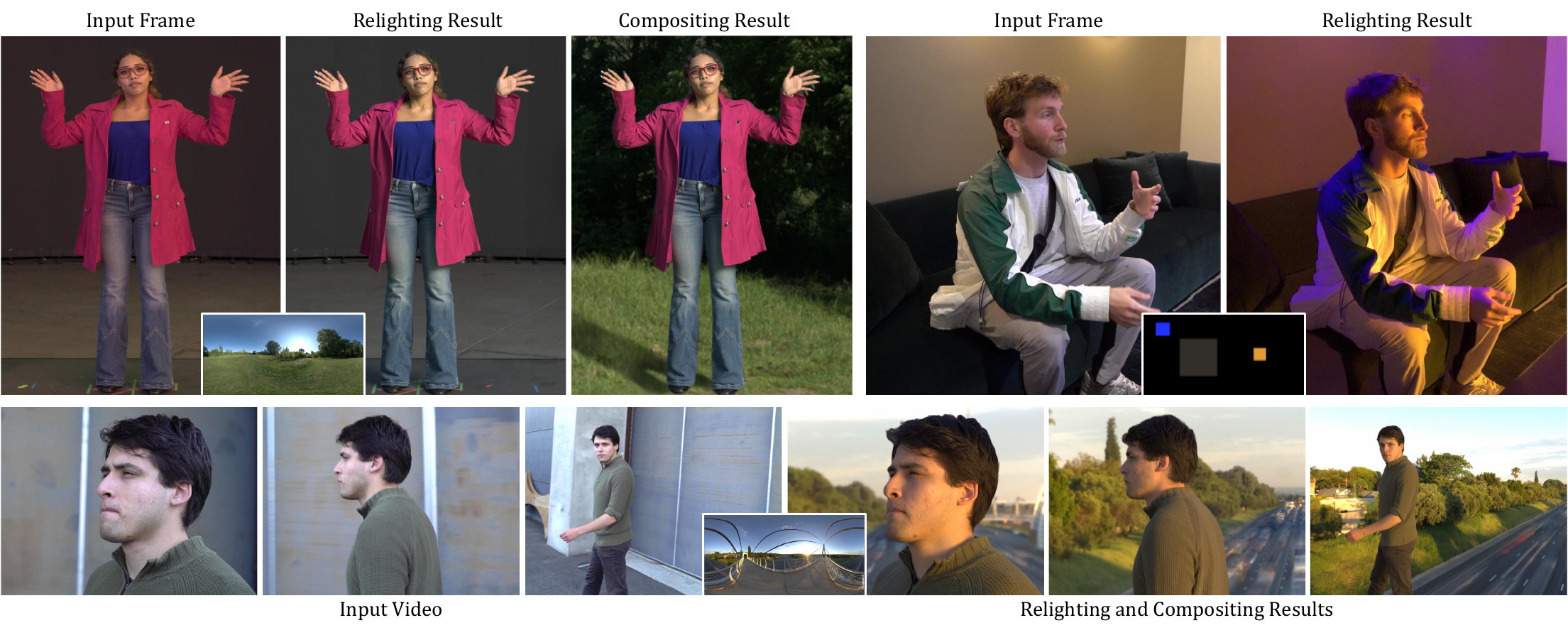}
\vspace{-8mm}
\caption{\textbf{Input and relighting results of \shortname.}  Given an input video of a subject with arbitrary lighting, and a target lighting condition, \shortname allows relighting the input video to the target lighting with a high level of photorealism and temporal consistency. 
The technique works for any framing (full body, upper body, and closeup), any resolution, and any frame length, including casually captured videos.
Compositing the relit results onto appropriate backgrounds ("compositing results") produces plausible complete shots. 
}
\label{fig:teaser}
\end{teaserfigure}

\maketitle

\raggedbottom

%%%%%% FOR ARXIV %%%%%%
% \pagebreak
%%%%%%%%%%%%%%%%%%%%%%%
\section{Introduction}

%%A. Relighting is important (and hard)

Lighting is fundamental to cinema, essential for underscoring atmosphere, emotion, and composition. 
But effective lighting is difficult to achieve: on set, a team of trained professionals must move stands, ladders, and lights, adjust dimmers, barn doors, and lenses, and must place filters and scrims and flags and bounce cards. 
On location, the cinematography often must wait for the sun to be in the right positions or for a cloud to pass. 
In every case, many other talented craftpeople must wait for the lighting to be ready so the next scene can be shot.
Being able to shoot scenes of a movie in whatever lighting is easily available, and, in post-production, changing the lighting to best serve the artistic storytelling needs of the scene, is highly appealing.
It could save a great deal of time and resources, but, more importantly, it would allow turning the task of lighting a shot as a malleable, decoupled and iterative task independent of budget, time and place constraints.
% it would allow the desired lighting to be achieved and tweaked regardless of the environmental conditions or time constraints of the day. 
% \AHMET{this may read better at the end..., it would allow turning the task of lighting a shot as a malleable, decoupled and iterative task independent of budget, time and place constraints.}

While appealing, production grade shot relighting is challenging. 
Many complex aspects of light transport such as shadows, specularities, interreflections, translucency, and scattering must be disentangled from the original scene's shape, material and performance, and simulated as under an entirely new lighting. 
Additionally, to satisfy the vision of the lighting designers, our method must be highly controllable.
% \AHMET{lighter is the more common term but lighting designer may be more generic perhaps} 
% Li: lighter might be too professional for academic reviewers to understand, I googled lighter and only thing I find is the flame generator.

%%Photorealistic video relighting for full body is hard

While recent models have shown significant promise for relighting faces~\cite{diffrelight,relightfulharmonization,luxpostfacto,SynthLight}, 
% \MINGMING{More related works?} 
relighting full-body performances in a controllable, realistic, and temporally stable way has remained elusive.
Firstly, human bodies are harder to relight than faces: they exhibit many more degrees of freedom in motion, leading to complex pose variations as well as significantly stronger self-shadowing and interreflections.
Moreover, full-body performances involve clothing, which introduces a vast range of material reflectance properties -- from glossy black leather to translucent white mesh -- whose appearance further changes as garments fold, stretch, and drape in complex ways during motion.
As a result, high-quality video data with sufficient diversity in both lighting and body poses is crucial for full-body performance relighting. However, capturing such video data remains difficult, as it requires recording multiple instances of the same video under different lighting conditions. Rapid switching between lightings often introduces strong strobing, causing subject discomfort.

%\li{Finally, acquiring video relighting data for training -- i.e., multiple instances of the same videos captured under different lightings -- is challenging.
%For image relighting, subjects can be instructed to hold as still as possible while the same pose is captured under different lightings using a light stage. For videos, however, switching between many lighting conditions typically introduces strong strobing effect and is uncomfortable for real-world use case.}
% Moreover, whereas a subject can be asked to perform a series of facial expressions and hold each pose long enough to be captured under multiple lighting conditions, this assumption does not hold for full-body performances.
% The space of possible body poses is far larger, and many poses are not statically balanced, making it impractical to capture a complete set of lighting conditions for diverse poses at scale. \li{should we claim this given that we also capture OLAT for body?}
% \TODO{add that if we capture video at two different light it will be flickering (w/o bipack).}

% Bodies are also clothed, so they can have essentially any material reflectance property from black shiny leather to translucent white mesh -- and these varying materials fold and stretch and drape in dramatically different ways as well. 
% And while a human subject can be asked to perform a series of facial expressions and hold each one long enough to record it in several lighting conditions, this is a far less reasonable request for full-body poses, since many more poses are possible, and many of them are not statically balanced.

To address these challenges, we introduce a novel framework for capturing and relighting full-body performances. We propose to acquire high-quality video relighting training data by combining static One-Light-at-a-Time (OLAT) images~\cite{firstlightstage} with dynamic sequences captured under slowly evolving paired lighting conditions. This enables both highly controllable illumination through OLATs and a wide range of body poses and facial expressions through temporally coherent video data, providing supervision for training a video relighting model. Moreover, for the dynamic sequences, we exploit digitally bi-packed lighting patterns~\cite{Yu:BiPack:2025} to produce paired video data without noticeable flicker to the subjects, enabling a comfortable capture environment.

%%\TODO{harmonize this paragraph}

%%Save for implementation section
%%

We then finetune a video diffusion model using the high-quality relighting data. 
The strong generative priors of the diffusion model help produce photorealistic and temporally consistent relighting results.
% The bi-packing yields several video clips of the subject under two different lighting configurations, which provides valuable supervision to finetune a video diffusion model; the strong priors help produce photorealistic and temporally consistent relighting results.  
To accurately condition the model on illumination, we introduce OLAToken, a new lighting-conditioning mechanism that learns a permutation-invariant aggregation of per-light contributions, closely reflecting the compositional nature of physical illumination.
In addition, we propose a dynamic lighting conditioning module that enables relighting under time-varying illumination, allowing lighting to change continuously throughout a performance.

To summarize, we first capture a subject under both static and dynamic poses with dynamic lighting patterns to produce paired video data with known illumination conditions.
We then train a subject-specific video relighting model that can relight arbitrary poses of the same subject under arbitrary lighting in a long sequence, producing high-resolution, temporally coherent, photorealistic relighting effects across the whole body as shown in Fig.~\ref{fig:teaser}.

%%\begin{itemize}
%%\item  Unlike face or static object relighting, body exhibit complex non-convex geometry with non-rigid motion, self-shadowing, clothing materials, etc. Therefore a high quality dataset
%%\item  Difficult to acquire \textbf{video} relighting data for temporal consistency.
%%\end{itemize}

% C. Exisiting works
% \begin{itemize}
% \item  intrinsic decomposition + PBR, struggle with photorealism (because limitation of PBR)
% \item  OLAT captures, but most are images, so lack temporal consistency. Also architecture wise stuggle to handle complex effects.
% \item  Scale up by using in the wild data, but quality depends on the pseudo pair, poor controllability (for background conditioning)
% \end{itemize}

% D. In this work
% \begin{itemize}
% \item  We use video diffusion model to achieve photorealistic and temporally consistent video relighting
% \item  to train this model requires diverse combination of lightings, performance sequences and view points. So we design new way to capture data that resulting in video relighting pairs.
% \item  new lighting conditioning algorithm OLAToken for accurate lightinging conditioning that provides a learned, permutation-invariant mechanism for aggregating per-light effects, closely matching the compositional nature of physical illumination. And dynamic lighting conditioning module that allows for changing lighting.
% \end{itemize}

Our key contributions can be summarized as follows:  
\begin{itemize}
    \item A diffusion-based video relighting model that achieves state-of-the-art photorealistic, controllable, and temporally consistent subject-specific performance relighting.
    \item \li{A novel lighting conditioning approach that supports both static and dynamic lighting control.}
    \item A comfortable capture process that avoids flickering lighting.
    \item The use of a hybrid dataset that captures both lighting and pose diversity and generates pixel-aligned video relighting pairs, with wide, medium, and close-up views of the subjects.
    %\item A sophisticated data augmentation pipeline  
    % \item A novel static and dynamic lighting conditioning algorithm.
\end{itemize}

% \section{Related Work}
% \subsection{Novel view synthesis}
% \subsection{Super Resolution}

\begin{figure*}
    \centering
    \includegraphics[width=\textwidth]{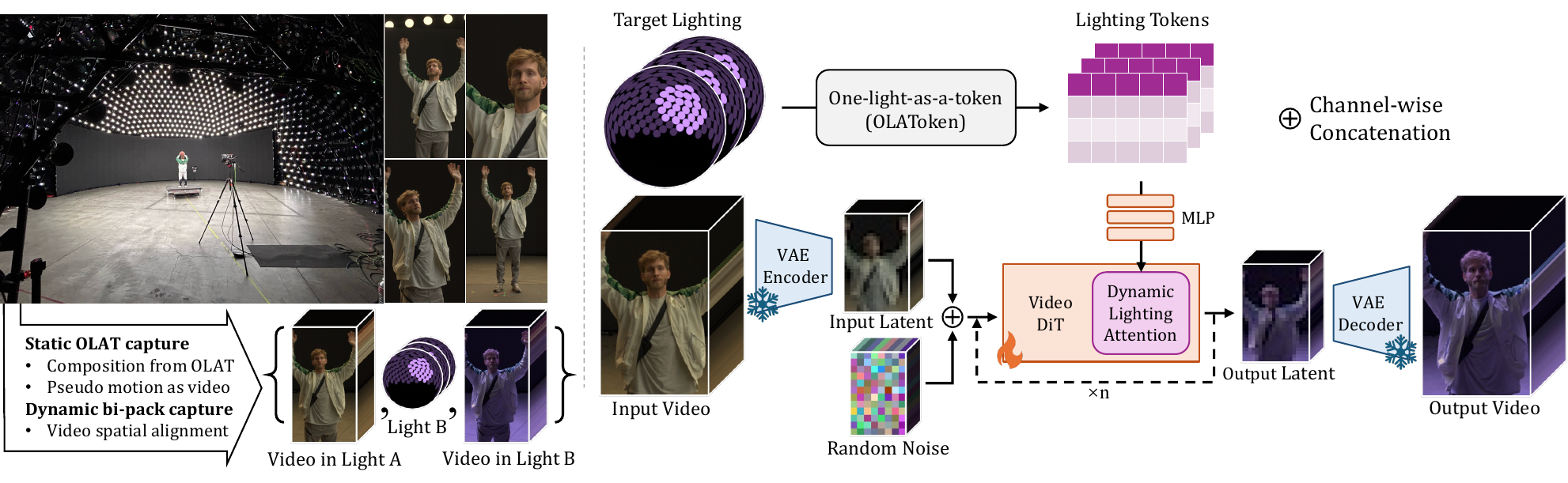}
    \vspace{-8mm}
    \caption{\textbf{Overview of the method.} We capture static OLAT data and bi-packed video data of a subject moving inside a large LED sphere, resulting in a dataset of video relighting training tuples that consists of two pixel-aligned videos under different lighting conditions and the corresponding lighting sequences. We train a video diffusion model with a novel lighting conditioning module that supports dynamic lighting control.
    }
    \vspace{-4mm}
    \label{fig:overview}
\end{figure*}

\section{Related Work}

\subsection{LED Spheres and Appearance Acquisition}

Sophisticated LED Spheres \cite{multispectral,lightingreproduction} have been proposed to give filmmakers controllable environmental lighting during shooting, but they do not offer the flexibility of relighting in postproduction. 
Relighting {\em after} shooting was addressed for static subjects with the first light stage 
\citet{firstlightstage}, which acquired the reflectance field of a human face using one-light-at-a-time (OLAT) sequences for image-based relighting. 

%%Alternative capture setups have been explored, using the sun  \cite{sunstage}, or synthetic data \cite{Lumos} as a substitute, however the quality is limited compared to a purpose-built light stage.

% By linearly combining OLAT images, any arbitrary lighting can be obtained.
%%For videos OLAT captures are impractical as the process is already asynchronous for static scenes.
%, thus introducing new challenges if we want to relight a dynamic sequences. 
Time-multiplexed illumination techniques \cite{time_multiplexed,locomotion} extend image-based relighting to dynamic subjects by looping though very rapidly repeating OLAT lighting patterns, but they require expensive ultra-high speed cameras (e.g. 1000+ fps) with limited spatial resolution, and tend to produce uncomfortable strobing illumination.
Optical flow \cite{reflectance_transfer,locomotion} and tracked meshes \cite{animatableface} can be used to propagate lighting across multiple relit reference images. 
While showing promise, this approach relies heavily on the quality of the alignment and struggles with highly dynamic content.
Recently, neural networks have been trained on OLAT reflectance data to implicitly learn a subject's appearance under different lighting \cite{deeprelightablemodel,deeprelightabletexture,diffrelight,deepreflectancefield} but have struggled with temporal stability.

% Removed because of space
% For the volumetric capture setup, several works explore various relightable 3D representations that are trained on OLAT captures, such as morphable mesh \cite{deeprelightablemodel}, morphable Gaussian Avatar \cite{body_gca,face_gca,uravatar,vrmm}, or OLAT Gaussian Splatting \cite{GS3,olatgaussian}.
% These methods either require some 3D proxy or volumetric capture of the performance to be relit or are limited to faces and temporally unstable.
% In contrast our method allows relighting any videos of the captured subject shot with the same costume.

\subsection{Physics-Based Relighting}

Instead of a purely data-driven approach, physics-based relighting recovers intrinsic or material attribute of images, and then use physics-based rendering (PBR) to relight them. 
Some early works infer a Cosine-based BRDF from color-gradient patterns \cite{cosinelobe,therelightables}. 
Retinex theory \cite{retinex} is used as a hand-crafted prior to separate reflectance and illumination. 
High quality intrinsic decomposition from a single image or videos has been achieve with deep learning \cite{rgb2x,firstintrinsic,irisformer,inverserenderandmcrt,DiffusionRenderer} by training on large scale synthetic data. 
With either hand-crafted or data-driven priors, many works use differentiable rendering to jointly optimize materials in 3D to get a relightable 3D representation \cite{nerv,invrender,physg,TensoIR,gsssr,svgir,gigs,gsir}. 

PBR based relighting often results in a synthetic looking appearance, especially for humans, whose skin and hair scatter light in complex ways. 
A neural renderer can be cascaded to PBR to fix these artifacts \cite{indoorneuralrelight,outcast,geometryawarenetwork,SwitchLight} but human appearance remains challenging.
Some recent works have shown that Diffusion Models have priors strong enough to bypass PBR and directly generate the final images from estimated materials and target lighting \cite{rgb2x,DiffusionRenderer}. 

\subsection{Deep Relighting}
% \AHMET{de facto is the right spelling but may be better to use the word common here, or the word prevailing}
Training a relighting model using a deep neural network has been a promising approach to achieving robust and generalizable relighting \cite{deepsingleimagerelight}. 
A common paradigm is to condition the model with intermediate buffers, such as gradient illumination \cite{deepreflectancefield}, normal and albedo \cite{totalrelighting,holorelighting,multiplereflectance,LightPainter,allfreq_fullbodyrelight}, roughness \cite{SwitchLight,allfreq_fullbodyrelight}, precomputed shading maps \cite{totalrelighting,dilightnet,SwitchLight,lightit,allfreq_fullbodyrelight,geometryawarenetwork} or shadow maps \cite{realisticshadow,lightit,allfreq_fullbodyrelight}. 
End-to-end prediction of the relighting results is also feasible if the model has a large capacity \cite{singleimagerelight,SwitchLight} or strong generative prior such as with a diffusion model \cite{diffrelight,iclight,luxpostfacto,lightlab,LBM,neuralgaffer,genlit}.
There is growing interest in using a large scale synthetic datasets to train relighting models \cite{Lumos,SynthLight,Multiilluminationsynthesis,iclight}.
Most of these methods either work on objects or single images and do not provide the quality required for professional video production.

% Regarding the lighting condition, HDRI is a commonly used lighting representation for human centric relighting. 

% Diffusion model has shown great results to generate

To train a video relighting model, NVPR \cite{nvpr} acquired video relighting data at 1000fps \citet{time_multiplexed}. 
To avoid using a high fps camera, Relumix \cite{relumix} and LightAVideo \cite{lightavideo} extend single image relighting models to video. 
\citet{Comprehensive_relighting} enforces temporal consistency by using flow-based warping. RelightVid \cite{relightvid} generates synthetic relighting pairs from in-the-wild videos through color augmentation. 
DiffusionRenderer \cite{DiffusionRenderer} and Unirelight \cite{unirelight} use synthetic 3D assets to train and tend to generate synthetic-looking results.  We believe real-world data is necessary for production grade relighting and propose a video capture process based on standard cinema cameras.

%Thus we propose a new way to capture video training data with a conventional cinema camera at 120fps.

% For full body relighting. 

\section{Method}

An overview of our pipeline is shown in Fig~\ref{fig:overview}. 
We first capture relighting data of the subject using a large LED Sphere. 
In addition to a set of traditional OLAT captures of static poses, we record paired video relighting training data using the digital bi-pack \cite{Yu:BiPack:2025} technique of interleaving two different lighting condition sequences.
After data preprocessing, we get a dataset of video training triples $\{V_A, L_B, V_B\}$, where $V_A$ and $V_B$ are videos with pixel-aligned content but with two different lightings, $L_A$ and $L_B$. 
From this dataset we train a video diffusion model for video-to-video translation that relights input videos to any target lighting condition. 

\subsection{Apparatus}

We employ a large-scale LED Sphere with approximately 
1600 custom LED light sources distributed over a spherical structure. Each light source has 216 high-power LEDs, spread across red, amber, green, blue, royal blue, and white. 
We use a real-time multi-spectral lighting reproduction algorithm \cite{RTmultispectral} to obtain high quality color rendition.
The lights switch to the next lighting condition with sub-microsecond level accuracy through a wired TTL pulse, allowing a lighting sequence pre-loaded into onboard flash memory to be played back in precise synchronization with the camera shutter as in \citet{time_multiplexed}.
\camready{The stage also includes a black LED panel wall. However, due to its different intensity and color rendition, as well as the lack of synchronization with the strobe system, it is not used in our capture process.}
% by maximizing the use of the broader spectrum amber and white LEDs in the color reproduction.

%%All lights are then able to simultaneously switch to a new lighting pattern more than a thousand times per second, making the real-time playback possible. This method bypasses the bottleneck of the \LightDome's communication bandwidth, making it possible for the high-speed time-multiplexed lighting application \cite{Yu:BiPack:2025}.

%%The \LightDome also provides external synchronization capacities through the GenLock protocol, facilitating the adoption of any industrial standard lighting, imaging and sound equipment.

\begin{figure}
    \centering
\includegraphics[width=\linewidth]{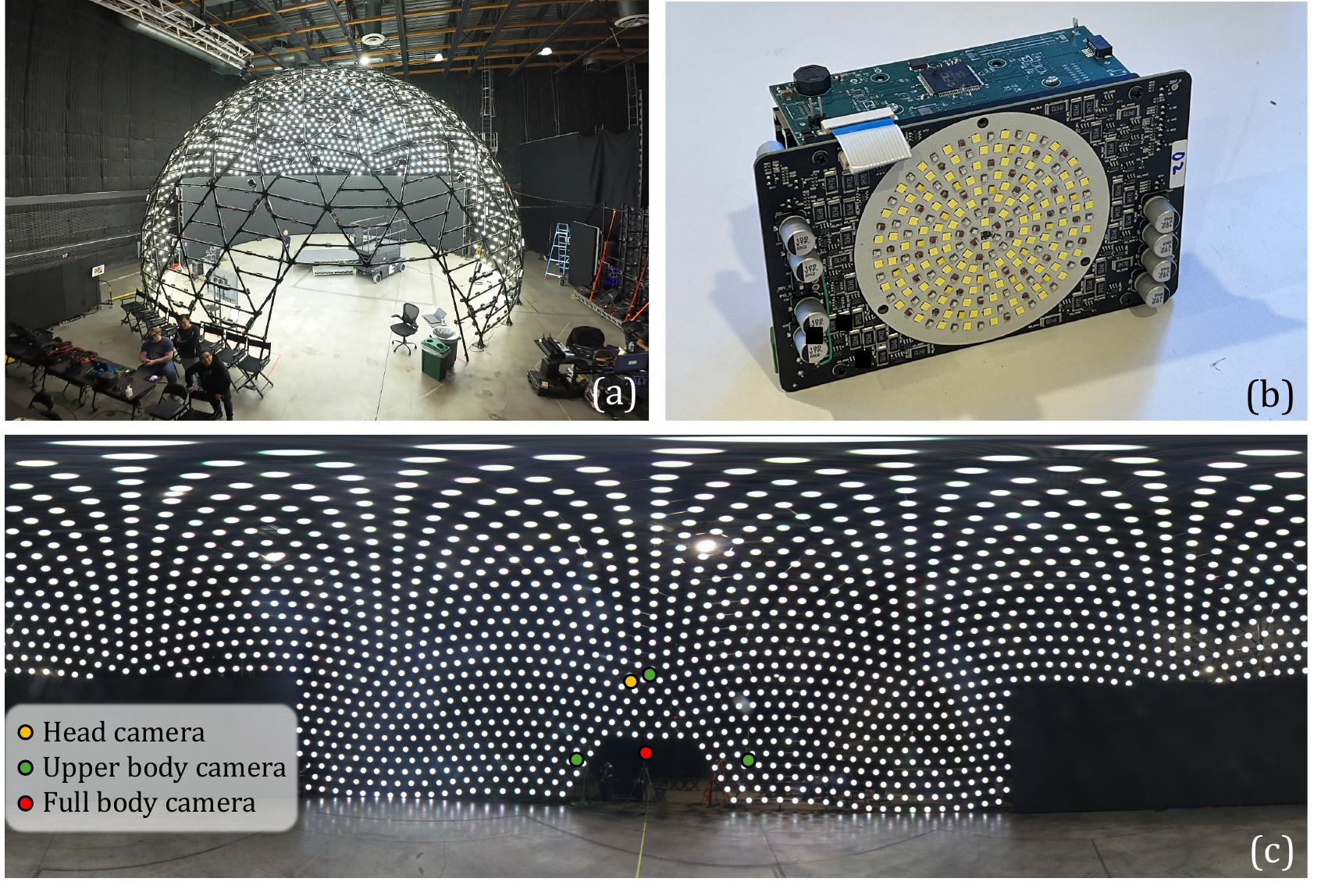}
\vspace{-6mm}
\caption{The \LightDome (a) consists of 1600 customized multi-spectrum lights (b). An equirectangular projection of all the lights can be seen in (c), where the distribution of cameras are visualized in colored dots.}
\vspace{-4mm}
\label{fig:hardware}
\end{figure}

We use five RED Komodo X cinema cameras placed near the periphery of the stage as in Fig.~\ref{fig:hardware}(c).  The cameras are outfitted with a combination of medium and long focal length lenses to produce wide, medium, and closeup shots of the subject.  The camera framed most tightly on the head uses a motorized pan-tilt system and optical tracking system to keep the subject's closeup in frame as they move.  The cameras are configured to capture at 120fps at UHD resolution (3840 $\times$ 2160), with a 360$^{\circ}$ shutter angle and f/4 aperture. 

%%%%%%%%%%%%%%%%%%%%%%%%%%%%%%%%%
%%%%%%%%%%%%%%%%%%%%%%%%%%%%%%%%%
%%%%%%%%%%%%%%%%%%%%%%%%%%%%%%%%%
%%%%%%%%%%%%%%%%%%%%%%%%%%%%%%%%%
%%%%%%%%%%%%%%%%%%%%%%%%%%%%%%%%%
%%%%%%%%%%%%%%%%%%%%%%%%%%%%%%%%%
%%%%%%%%%%%%%%%%%%%%%%%%%%%%%%%%%
%%%%%%%%%%%%%%%%%%%%%%%%%%%%%%%%%

\subsection{Capture process}
% emphasize the requirements
Our capture process is designed to comfortably record rich lighting data to train a subject-specific relighting model with the following criteria:

%%This requires these criteria \AHMET{This requires the following may read better here}:

\begin{enumerate}[topsep=0pt]
    \item Cover a varied set of static and dynamic lighting conditions
    \label{enum:light}
    \item Cover diverse view points and performances
    \label{enum:viewpose}
    \item Contain video training pairs under two well-synchronized lighting conditions
    \label{enum:video}
    \item Avoid flickering light perception so the subjects remain comfortable
    \label{enum:flicker}
\end{enumerate}

The biggest challenge is to design lighting patterns from which we can extract video training pairs (\ref{enum:video}), while at the same time guaranteeing the diversity of lighting patterns (\ref{enum:light}). 
%We choose to use an HDRI instead of OLAT as our control signal since it relieved from running multiple inference for OLAT and then compositing in post process \cite{deepreflectancefield,diffrelight}.
Most existing works \cite{time_multiplexed,locomotion,nvpr} capture time-multiplexed videos using high-fps ($>$1000fps) cameras so that multiple interleaved lighting conditions can be recorded closely together in time.

\begin{figure}
    \centering
    \includegraphics[width=\linewidth]{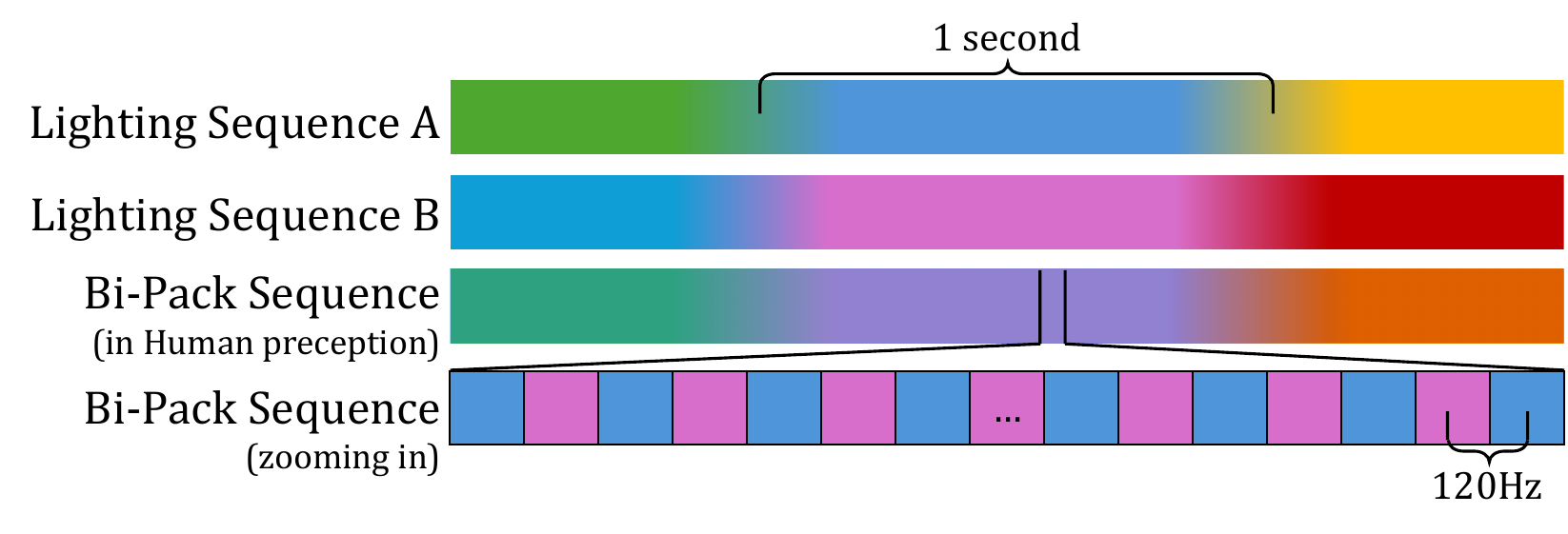}
    \vspace{-6mm}
    \caption{\textbf{Visualization of bi-pack lighting sequence.} \li{A bi-pack sequence consists of two lighting sequences that vary smoothly every 1 second, while rapidly alternating between the two at 120 Hz. Because the switching frequency exceeds the human flicker-fusion threshold, it appears as a mixture of two lightings evolving at 1 Hz.}}
    % \TODO{Note from JP: Shouldn't the first and second line have alternating empty(black) light, I find it confusing here}
    \vspace{-4mm}
    \label{fig:bipack}
\end{figure}

To use standard cinema cameras, we record each subject performing a few-minute of movement sequences at 120 fps.
As subjects perform, the LED Sphere digitally bi-packs \cite{Yu:BiPack:2025} two gradually transforming spherical environment lighting sequences, A and B, each running at 60 fps.
This keeps high-frequency lighting changes above the human visual system's 60Hz flicker fusion frequency for the comfort of the subject as shown in Fig.~\ref{fig:bipack}.  
After temporally aligning sequences A and B, we have a plethora of paired video clips of the subject performing the same aligned motions but recorded under two different lighting conditions. 

%%In our case, with 120fps cameras, we can only afford squeezing (which we call bipack) 2 lighting conditions based on the flicker constraints (\ref{enum:flicker}). In order to meet the lighting diversity requirements (\ref{enum:light}), we gradually fade the lighting to new conditions about every second as the subjects perform.  The resulting lighting pattern is visualized in Fig.~\ref{fig:bipack}.  Because two lighting patterns are still time multiplexed at 120Hz, human eye will percept the lighting as a mixture of two lighting patterns smoothly changing at 1Hz.

Since each lighting sequence evolves slowly, with smooth changes happening around once per second, the number of unique lighting conditions is limited. Therefore, we also capture OLATs of a small set of static poses \cite{firstlightstage,diffrelight}. 

In summary, our capture process includes both static OLAT captures and dynamic bi-packed captures.
We predefined 13 static poses and 9 dynamic performance sequences and ask the subject to perform accordingly. 
To get a diverse set of view points (\ref{enum:viewpose}), we asked the subject to turn to the 4 cardinal directions in each take. 
Each take lasts around 40 seconds for a static OLAT capture and 70 seconds for a dynamic bi-pack capture, with a total capture time of 30 minutes per subject.
The details of the predefined poses and lighting designs are presented in the supplementary material. 

%%%%%%%%%%%%%%%%%%%%%%%%%%%%%%%%%
%%%%%%%%%%%%%%%%%%%%%%%%%%%%%%%%%
%%%%%%%%%%%%%%%%%%%%%%%%%%%%%%%%%
%%%%%%%%%%%%%%%%%%%%%%%%%%%%%%%%%
%%%%%%%%%%%%%%%%%%%%%%%%%%%%%%%%%
%%%%%%%%%%%%%%%%%%%%%%%%%%%%%%%%%
%%%%%%%%%%%%%%%%%%%%%%%%%%%%%%%%%
%%%%%%%%%%%%%%%%%%%%%%%%%%%%%%%%%

\subsection{Data Preprocessing}

After capture, we get a set of video files from each camera.
We extract linear EXRs and their corresponding lighting configurations. 
We preprocess this data to extract video relighting training tuples, $\{V_A, L_B, V_B\}$, where $V_B$ represents a video sequence under lighting $L_B$, and $V_A$ and $V_B$ are pixel aligned and differ only in lighting.

For static OLAT capture, we align all the OLATs to a reference tracking frame as in \cite{diffrelight,time_multiplexed} to obtain pixel-aligned OLAT sequences. 
We then generate HDRI-lit images by compositing OLAT sequences in linear color space \cite{firstlightstage}. 
For each take, we randomly select 50 HDRIs from Polyhaven \cite{polyhaven_hdri} with a random horizontal rotation to create a dataset with diverse lighting.
To get video pairs, we repeat the image along the time axis while adding translation and zoom motion.

For dynamic Bi-Pack captures, we extract two video sequences at 60fps with different lightings, noted as $V_A$ and $V_B$. 
Since the corresponding frames in the two videos are still captured at slightly different times, they suffer from small misalignments, which could cause ambiguity while training. 
Therefore, for every two consecutive frames of the same lighting condition, we interpolate a middle frame using FILM \cite{film,film-tf}. 
This results in two 120fps interpolated videos, $V_A'$ and $V_B'$ where the frames of $V_A'$ are temporally aligned with those from $V_B$, and those from $V_B'$ with $V_A$. 
Since these processed frames may exhibit artifacts, we only use them as conditioning to prevent our relighting model from learning to synthesize artifacts. 
Formally, we obtain two training pairs $(V_A', B, V_B)$ and $(V_B', A, V_A)$.

%%%%%%%%%%%%%%%%%%%%%%%%%%%%%%%%%
%%%%%%%%%%%%%%%%%%%%%%%%%%%%%%%%%
%%%%%%%%%%%%%%%%%%%%%%%%%%%%%%%%%
%%%%%%%%%%%%%%%%%%%%%%%%%%%%%%%%%
%%%%%%%%%%%%%%%%%%%%%%%%%%%%%%%%%
%%%%%%%%%%%%%%%%%%%%%%%%%%%%%%%%%
%%%%%%%%%%%%%%%%%%%%%%%%%%%%%%%%%
%%%%%%%%%%%%%%%%%%%%%%%%%%%%%%%%%

\begin{figure*}[t]
    \centering
    \includegraphics[width=\linewidth]{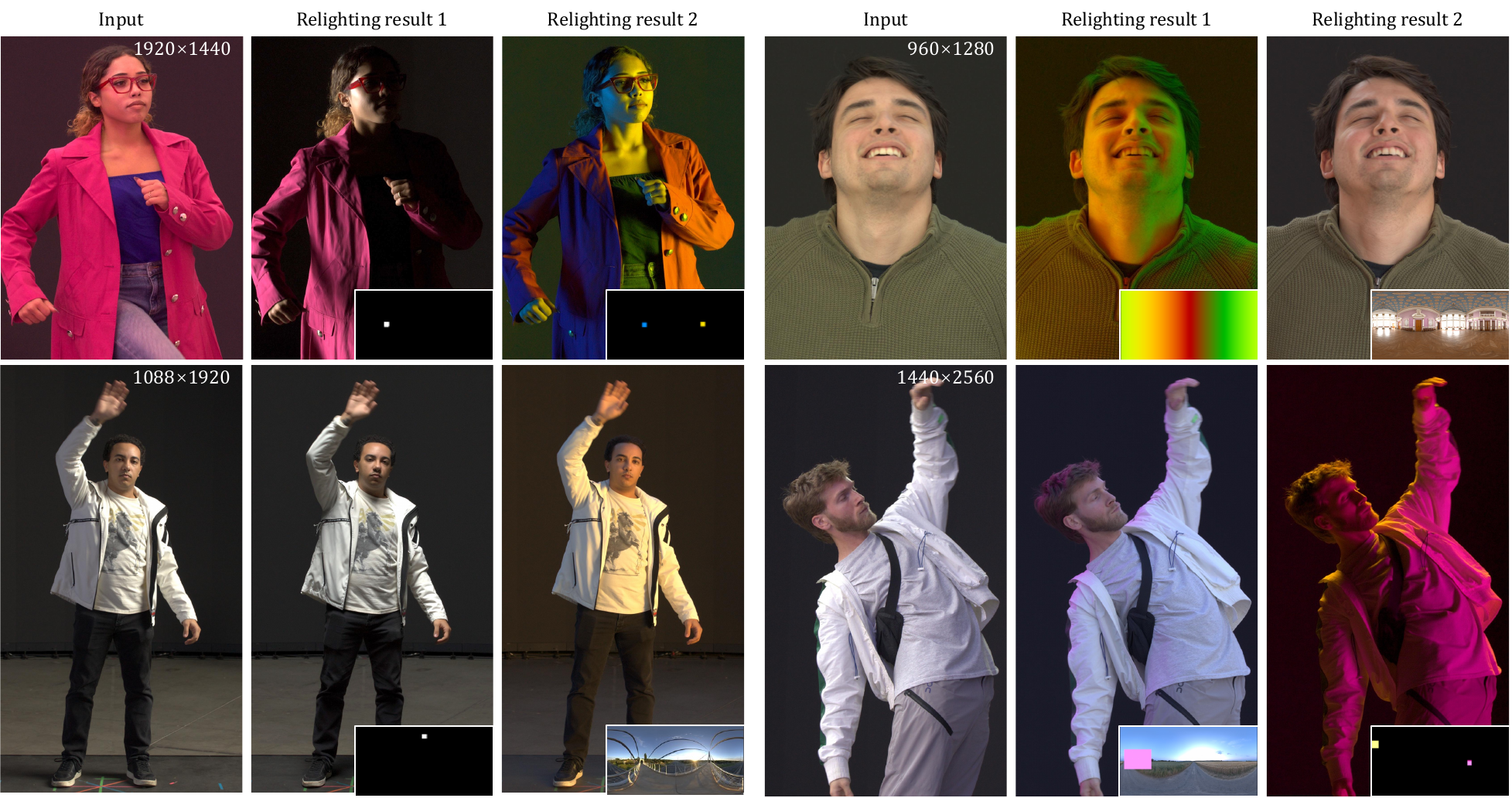}
    \vspace{-6mm}
    \caption{\textbf{Relighting results.} We show input videos and relighting results under novel performance and lighting conditions. Each result is a single from from a relit video. Our method achieves photorealistic relighting under directional lighting, multiple point lights, gradient illumination, image-based lighting, and manually edited HDRI maps for various input aspect ratios and resolutions. The resolution is shown on the top right and the lighting condition is visualized on the bottom right as a lat-long map.}
    \vspace{-4mm}
    \label{fig:mainresult_relight}
\end{figure*}

\subsection{Video Relighting Models}

Given a dataset of video training tuples $\{(V, L, V_L)\}$, we train a subject-specific video relighting model to relight video from any input lighting to any desired lighting. 

\paragraph{Preliminary}
For a practical capture process, our data covers just a sparse subset of viewpoints, motions, and lighting conditions.
We thus require strong video priors to interpolate this space and leverage a pretrained video diffusion model, WAN2.2 5B \citet{wan2025}, as the backbone of our video relighting model.
%First, a video variational autoencoder (VAE) compresses the input video $V$ into a latent $z \in \mathbb{R}^{h\times w\times t\times c}$.
This model takes a text prompt and a random latent $z^{(T)} \in \mathbb{R}^{h\times w\times t\times c}$ and flattens it into a list of tokens.
Then a Diffusion Transformer (DiT) iteratively de-noises the noisy latent to a clean one $z^{(0)}$, following the reversed flow matching process \cite{flowmatching}.
The DiT is a cascade of transformer layers to which text tokens are cross-attentioned to achieve text conditioned generation.
Finally the decoder of a variational autoencoder (VAE) is used to obtain an RGB video.

\paragraph{Spatial Conditioning}
To achieve spatial conditioning of the input video $V$, inspired by several diffusion-based pixel-to-pixel translation frameworks \cite{brooks2023instructpix2pix,ke2023repurposing,diffrelight,iclight}, we concatenate the conditioning latent $z_V$ with the noisy latent $z^{(t)}$ along the channel dimension, and duplicate the input channel number in the first \li{convolution} layer of the patchifier.
We find this simple approach effectively preserves the details, while introducing minimal overhead to both training and inference. 
% \TODO{JP: is this a convolution or linear layer? (confirmed it's convolution)} 

% dynamic lighting conditing
\paragraph{One Light as a Token}  \hspace{3mm} % why is this here and not with any of the other paragraphs?
Conditioning the model on the lighting $L$ during diffusion introduces new challenges.
To avoid generating hundreds of OLAT videos to be composed \cite{diffrelight}, we directly condition the model on a list of light sources to relight with a single inference. 
Existing works have explored the use of pixel-aligned shading maps rendered from an HDRI and estimated geometry \cite{totalrelighting,dilightnet}, but these are sensitive to errors and flickering, common with video data.
Some encode HDRIs into lighting embeddings using either dedicated lighting encoder \cite{iclight,luxpostfacto} or pretrained image VAE \cite{SynthLight,DiffusionRenderer,unirelight}. 
These methods effectively compress the lighting along the channel dimensions, leading to information loss and inaccurate controls.
% Some first encode HDRIs into single or few light embeddings \cite{iclight,luxpostfacto}, where all light sources are mixed and blended into the channel dimensions. 
% \li{This ``compression'' of the light sources could leads to information loss, resulting in inaccurate conditioning.}
% \TODO{JP:what is the issue with this approach? this sentence must lead to a limitation}
% Finally, using the pretrained VAE to encode the HDRI , suffers from information loss when encoding high dynamic range content as the VAEs are trained to encode 8-bit content. 

A good lighting condition should preserve the full dynamic range and directionality. 
Therefore, we introduce \textbf{OLAToken}, which stands for One-Light-as-a-Token. 
We treat each light source as one light embedding containing its intensity and direction in their own channels.
Each light token is passed through a small MLP to match the channel dimension of the model, and is fed into the DiT through cross-attention. 
% \TODO{JP: How is that different from LuxPostFacto?}
This simple design introduces two strong inductive biases that approximate the physical structure of light transport: 
(1) \textit{Permutation invariance}. 
Similar to a list of lights, the order of the tokens does not affect the final results in cross-attention.
And (2) \textit{Compositionality}. 
Although cross-attention is not inherently linear as lighting should be, it behaves as a context-dependent weighted summation, which allows the model to approximate the physical superposition of illumination.
This approach is more general than the lighting encoder used in \citet{luxpostfacto}. It allows flexible combinations of light sources from different directions, without the constraints of fixed resolution and orientation.

% Formally, given a target HDRI,
Formally,a target lighting condition is represented as a list of light sources $L = \{(\mathbf{l}_i, \mathbf{d}_i)\}$, where $\mathbf{l}_i \in \mathbb{R}^3$ is the linear intensity of the RGB channels, and $\mathbf{d}_i$ is the light direction in camera space.
For efficiency, we downsample the lights by evenly sampling $K$ directions over the hemisphere, and then aggregate every light source in $L$ to the closest points to form a new light source $(\mathbf{I}_j, \mathbf{D}_j)$:
\begin{equation}
    \mathbf{I}_j = \sum_{i\in \mathbb{J}} \mathbf{l}_i \text{, and } \mathbf{D}_j = normalize({\frac{\sum_{i\in \mathbb{J}} \mathbf{d}_i \|\mathbf{l}_i\|_2}{\sum_{i\in \mathbb{J}} \|\mathbf{l}_i\|_2})},
\end{equation}
where $\mathbb{J}$ is a set of indices for all the closest light sources close to $j$-th point. 
Then an OLAToken $\mathbf{T}_j$ is computed as:
\begin{equation}
    \mathbf{T}_j = \mathcal{D}(\mathbf{D}_j) \oplus \mathcal{I}(\mathbf{I}_j).
\end{equation}
Here, $\mathcal{D}$ is a directional encoding represented by Fourier Features similar to NeRF \cite{nerf}.
% \TODO{JP: How do we use SH to encode a single direction? This is not typical} 
$\mathcal{I}$ is a color encoding function.
We use a series of different gamma functions (log spacing from $1/3$ to $3$) to enhance the expressiveness of the color intensity. 
$\oplus$ indicates channel concatenation.

\begin{figure*}
    \centering
    \includegraphics[width=\linewidth]{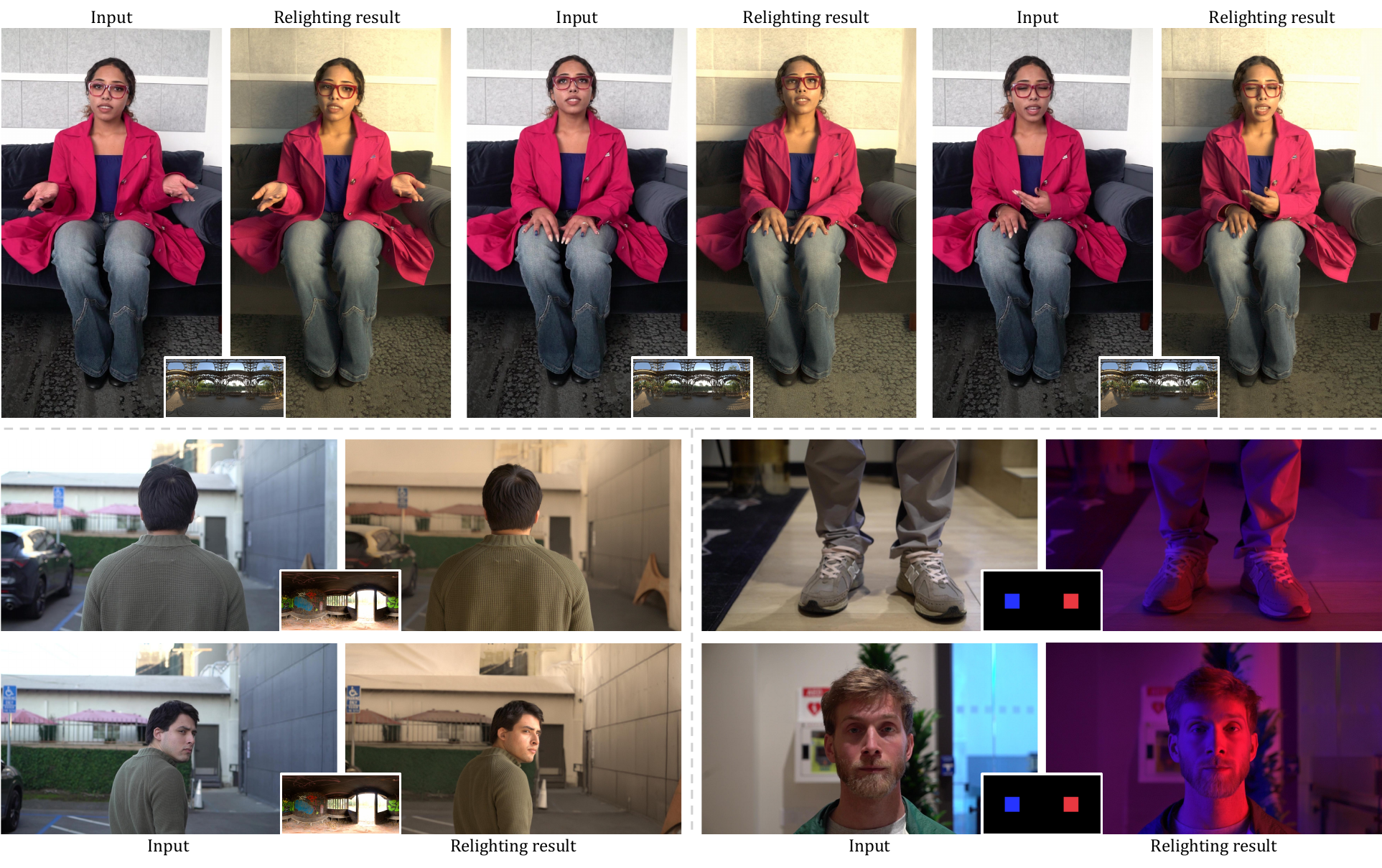}
    \vspace{-8mm}
    \caption{\textbf{In-the-wild relighting results.} We show video relighting results for in-the-wild captures.}
    \vspace{-2mm}
    \label{fig:mainresult_inthewild}
\end{figure*}

\paragraph{Dynamic lighting attention}
Importantly, we want the ability to condition the video relighting model on dynamic lighting where the lighting condition is specified for each frame, such as an environment rotating around the subject, or the subject walking into sunlight.
We show in the experiments that when adding time embeddings to the OLATokens, conditions of specific frames tend to leak into other frames. 
To address this, we use block diagonal attention masks during cross-attention such that the tokens from a specific frame only attend to the light tokens corresponding to that frame.
In practice, to avoid computation where the mask is empty, we flatten the temporal dimension with the batch dimension to achieve frame-wise cross-attention.
% we permute the dimensions latent and light tokens such that the temporal axis are flattened with the batch dimension.
% \TODO{JP: what is said before is unclear. Typo?}

\paragraph{Loss and training}
With the input video condition and light condition, we convert the video generation backbone into a video relighting model.
At training time, we freeze the VAE encoder and decoder, and finetune only the video diffusion DiT. 
We use a standard flow-matching training loss:
\begin{equation}
    \mathcal{L} = \| \hat{\mathbf{\epsilon}}(\mathbf{z}^{(t)}; \mathcal{D}(V), L, t) - \mathbf{\epsilon}
    \|_2,
\end{equation}
where $\mathbf{z}^{(t)}$ is the noisy latent at time step $t$ and epsilon is the added noise. 
% \TODO{JP: Do we actually predict the noise, or the velocity as with FLUX? Confirmed and it's predicting the noise}
$\hat{\mathbf{\epsilon}}$ indicates the noise predicted by the DiT model, conditioned on the input latent $\mathcal{D}(V)$, lighting $L$ and time step $t$. 
% To emphasize foreground and saliency part such as human hand and head, we weighted the loss using a spatially varying weighted map.
% The weight map is converted from a segmentation mask predicted by Sapien model \cite{sapiens}. We set the weight of background to 0.1, the head and hand to 4, and other human parts as 1. 

To support arbitrary aspect ratios and frame length during training, we randomly select an arbitrary aspect ratio between $1:2$ and $2:1$, and a sequence length from $9$ to $37$.
We then determine the spatial resolution by limiting the number of tokens to $5$k and randomly crop the input video.
We achieve frame rate augmentation by retiming using the nearest neighbor.
By applying these data augmentations, we obtain a video relighting model that is robust to different aspect ratios, framing, resolutions, and frame rates.

We implement our model using DiffSynth \cite{diffsynth} based on the WAN2.2 5B model \cite{wan2025}. 
We train our relighting model on 8 NVIDIA A100 GPUs with 80GB memory each, for 100K iterations with a batch size of 8 taking around 3 days. 
We use the Adam Optimizer with a learning rate of 4e-5.

\paragraph{Long video inference}
While our model is trained on videos that have at most $37$ frames, we empirically find the model is robust to videos with up to $100$ frames.
To infer on videos of unconstrained length, we apply the relighting model on overlapping windows and utilize MultiDiffusion \cite{multidiffusion} to combine the predictions into temporally consistent video results.

\begin{figure*}
    \centering
    \includegraphics[width=\linewidth]{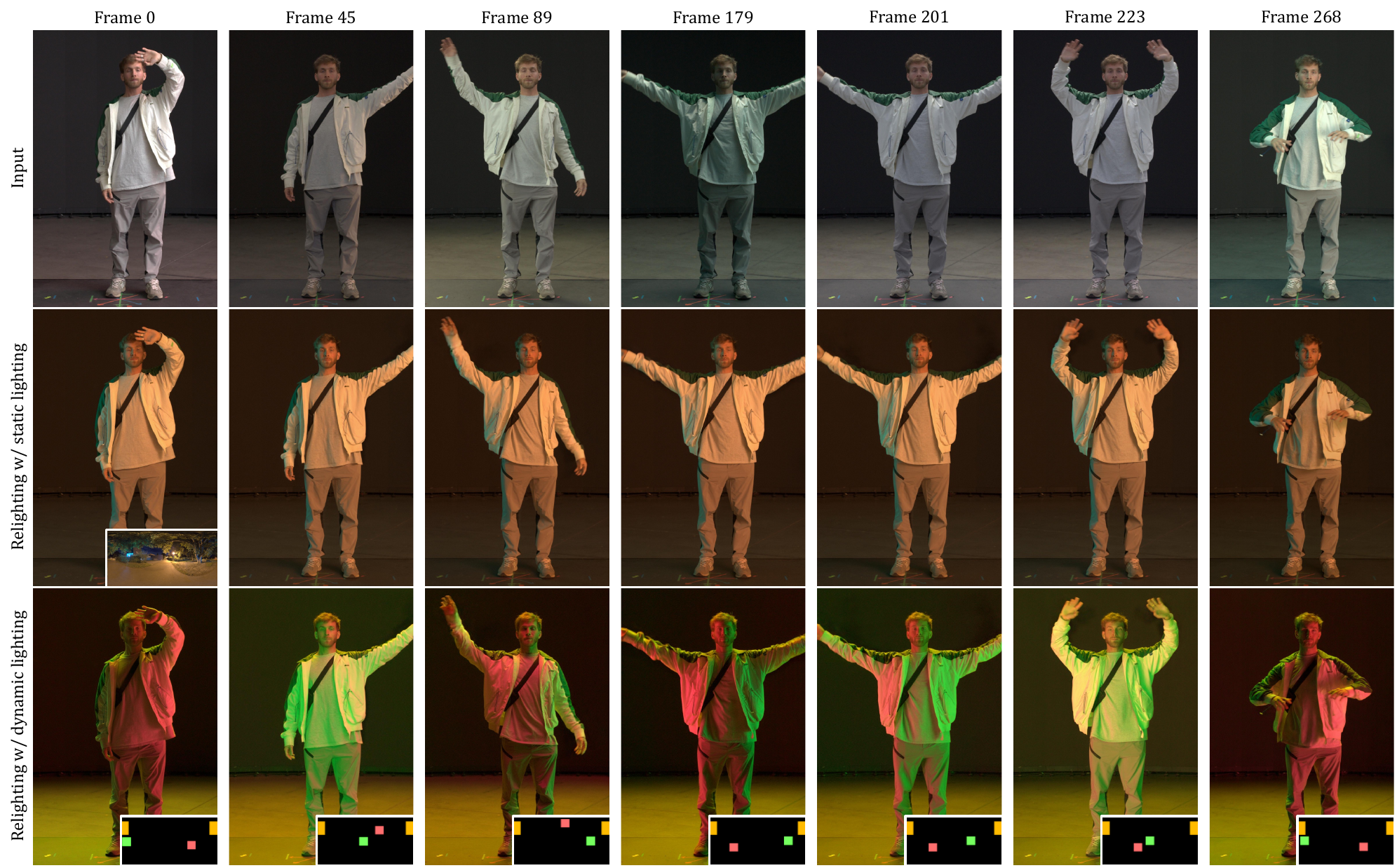}
    \vspace{-5mm}
    \caption{\textbf{Video Relighting results.} We show video relighting results of \shortname under static lighting and dynamic lighting conditions. Note that our method also works for input video with dynamic lighting conditions, while still being able to produce consistent relighting results. }
    \label{fig:mainresult_video}
    % \vspace{-1mm}
\end{figure*}

\section{Experiments}

We conduct several experiments to evaluate the performance of \shortname. 
We capture static OLAT and bi-packed video relighting data from four performers and train a single subject-specific relighting model on all four subjects. 
%%To evaluate the relighting, we also record several video clips per subject with new performances and lighting conditions. 
We also hold out several bi-packed lighting sequences from the training data to produce input videos with the same performance in a ground-truth alternate lighting condition, which we use to evaluate the relighting results both qualitatively and quantitatively.
During training, we hold out one camera view to ensure that the evaluation sequences include novel viewpoints.

For qualitative evaluation, we additionally capture several dynamic sequences for each subject in in-the-wild settings using a Sony Alpha $\alpha$7S at 50 fps with auto exposure enabled. We record performances in a variety of natural environments and ask each subject to perform freely as in a real-world scenario.

\subsection{Main Results}

We demonstrate the effectiveness of our method on several test sequences. 
Example frames are shown in Fig.~\ref{fig:teaser} and Fig.~\ref{fig:mainresult_relight}. 
Thanks to a unified light representation using OLAToken, \shortname achieves photorealistic relighting under a wide range of lighting conditions, including single and multiple point lights, gradient illumination, HDRI, and manually edited HDRI. 
The method is also robust to input videos with varying aspect ratios and resolutions. 

Qualitative video frames are shown in Fig.~\ref{fig:mainresult_video}; we refer the readers to the supplementary video for dynamic relighting examples. 
Using our long-video inference strategy, our method can process videos exceeding 200 frames. 
Because of the dynamic lighting conditioning, \shortname can relight sequences under smoothly varying lighting conditions that are not observed during training. 
Notably, when the input video itself contains changing illumination, our method still produces temporally consistent relighting results.

Moreover, although trained only on stage-captured data, our model generalizes well to in-the-wild captures, as shown in Fig.~\ref{fig:teaser} and Fig.~\ref{fig:mainresult_inthewild}. 
We evaluate several challenging scenarios that are unseen during training, including diverse performances, unusual framings, and camera motions. 
Our method achieves photorealistic quality across almost all the examples. 
Although our model is trained only to relight the subjects, we find that as a by-product it often relights the backgrounds somewhat plausibly. 
We attribute this effect to the priors of the base model, which is trained to generate plausible videos.
%% Background relighting should be regarded as a by-product rather than the focus of this work, as our primary goal is to enable realistic compositing of relit performances.

\begin{table}[t]
\centering

\caption{\textbf{Quantitative comparisons.} Ours achieve best relighting results while being the most temporally consistent. We also measure the inference time per frame on one A100 machine and ours achieve a reasonable speed.}
\vspace{-3mm}
\label{tab:comparison}
\small
\begin{tabular}{lccccc}
\toprule
name & PSNR$\uparrow$ & SSIM$\uparrow$ & LPIPS$\downarrow$ & T-PSNR$\uparrow$  & Inf. time$\downarrow$ \\
\midrule
Ours & \textbf{23.43} & \textbf{0.9441} & \textbf{0.04208} & \textbf{27.90} & 1.8s \\
DiffRelight+ & 21.75 & 0.9425 & 0.08095 & 26.34 & 4.2min\\
Switchlight3 & 16.88 & 0.8814 & 0.11945 & 26.84 & 1.7s \\
LuxPostFacto & 12.84 & 0.8407 & 0.13341 & 27.18 & 8.7s \\
Allfreq & 11.85 & 0.8139 & 0.20248 & 21.95 & \textbf{0.72s} \\
\bottomrule
\end{tabular}
\end{table}

\subsection{Comparisons}

\begin{figure*}
    \centering
    \includegraphics[width=\linewidth]{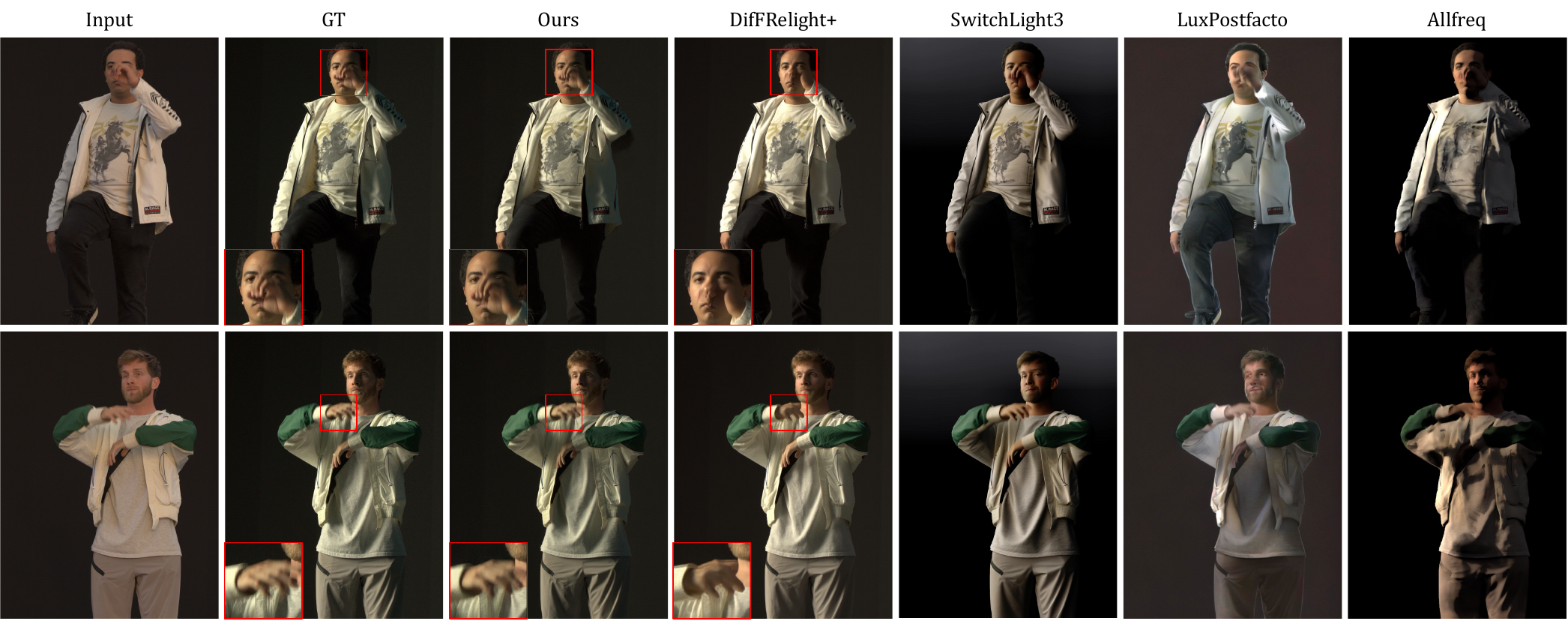}
    \caption{\textbf{Qualitative comparisons.} We compare ground truth and predicted relighting results. Ours achieves the most photorealistic relighting results. DifFRelight+ achieves reasonable results, but tends to produce artifacts in fast-moving regions, as highlighted in red boxes. Other generalized relighting models produce larger errors compared to the ground truth.}
    % \MINGMING{SwitchLight->SwitchLight3}
    \label{fig:comparison}
\end{figure*}

We compare our method with several baselines.
Closest to ours is DifFRelight \cite{diffrelight}, an image diffusion-based, subject-specific relighting model for facial performances.
We re-implement DifFRelight using our network backbone and retrain it on our data to align the experiment settings; this baseline is denoted as DifFRelight+.
We also compare against generalized human-centric relighting models: Switchlight3 \cite{SwitchLight}, LuxPostFacto \cite{luxpostfacto}, and Allfreq \cite{allfreq_fullbodyrelight}.
Since not all methods support dynamic lighting, we evaluate all approaches using clips with constant lighting conditions.
We report PSNR, SSIM, and LPIPS \cite{lpips} between relighting results and ground truth in Tab.~\ref{tab:comparison}. As our focus is human-centric relighting, we use the Sapiens model \cite{sapiens} to extract human segmentation masks, and compute all metrics only within the masked regions. 
We measure temporal consistency using T-PSNR, which is defined as the PSNR between the current frame and its optical-flow–warped neighboring frame.
We also report per-frame inference time at 1080p resolution.
The quantitative results show that our method achieves the highest relighting quality among all baselines, along with the best temporal consistency, while remaining reasonably efficient. 
DifFRelight+ performs slightly worse due to the lack of video-based training and is significantly slower at inference, as it requires running the model hundreds of times to generate all OLATs followed by post-compositing. 
Overall, subject-specific methods substantially outperform generalized models, highlighting the importance of subject-specific training for high-quality relighting.

We further present qualitative comparisons in Fig.~\ref{fig:comparison}. 
Our method produces photorealistic relighting results that are closest to the ground truth. 
In comparison, DifFRelight+ achieves reasonable quality but often introduces artifacts in ambiguous regions such as occlusions and fast-moving parts. 
This limitation arises because DifFRelight+ operates on single frames. 
SwitchLight3 produces plausible relighting but generates overly glossy and unnatural appearances, particularly on skin regions. 
This is likely caused by the physically based rendering priors it is using, which do not capture the complex reflectance of human skin. 
Other generalized relighting methods, such as LuxPostFacto, achieve plausible temporal consistency but exhibit pronounced artifacts and significantly alter subject identity due to the lack of subject-specific training data.

\subsection{Ablation Studies}
We conduct ablation studies to evaluate key design choices in our method. 
Experiments are performed on all evaluation sequences from the four subjects, which include novel performances, camera viewpoints, and lighting conditions. 
The test set consists of 50\% sequences with constant lighting and 50\% with dynamic lighting. 
A summary of the quantitative results is provided in Tab.~\ref{tab:ablation}. \li{Additional experiments are provided in the supplementary material.}

\begin{table}[t]
\centering
\small
\caption{\textbf{Quantitative ablation results.} Showing the effectiveness of static OLAT, bi-pack video data, as well as the lighting conditioning method. The best and second-best results are highlighted in \textbf{bold} and \underline{underline}, respectively. }
\label{tab:ablation}
\vspace{-3mm}
\begin{tabular}{lccc}
\toprule
 & PSNR$\uparrow$ & SSIM$\uparrow$ & LPIPS$\downarrow$ \\
\midrule
Full & \underline{22.62} & \underline{0.9369} & \textbf{0.04895} \\
w/o video data & 21.50 & 0.9338 & 0.07463 \\
w/o OLAT data & 22.03 & 0.9275 & 0.05685 \\
% w/o alignment & 21.45 & 0.9260 & 0.05692 \\
w/o OLAToken & 20.39 & 0.9312 & 0.08011 \\
w/o dyn. cond. & 22.31 & 0.9368 & 0.04913 \\
% w/o data aug. & 21.90 & 0.9330 & 0.05965 \\
% w/ random init. & & & \\
w/o alignment & 21.45 & 0.9260 & 0.05692 \\
w/o pretrained weight & 17.69 & 0.9055 & 0.09099\\
\camready{w/ WAN2.1 1.3B} & \camready{\textbf{22.88}} & \camready{\textbf{0.9388}} & \camready{\underline{0.05011}} \\

\bottomrule
\vspace{-4mm}
\end{tabular}
\end{table}

\begin{figure}
    \centering
    \includegraphics[width=\linewidth]{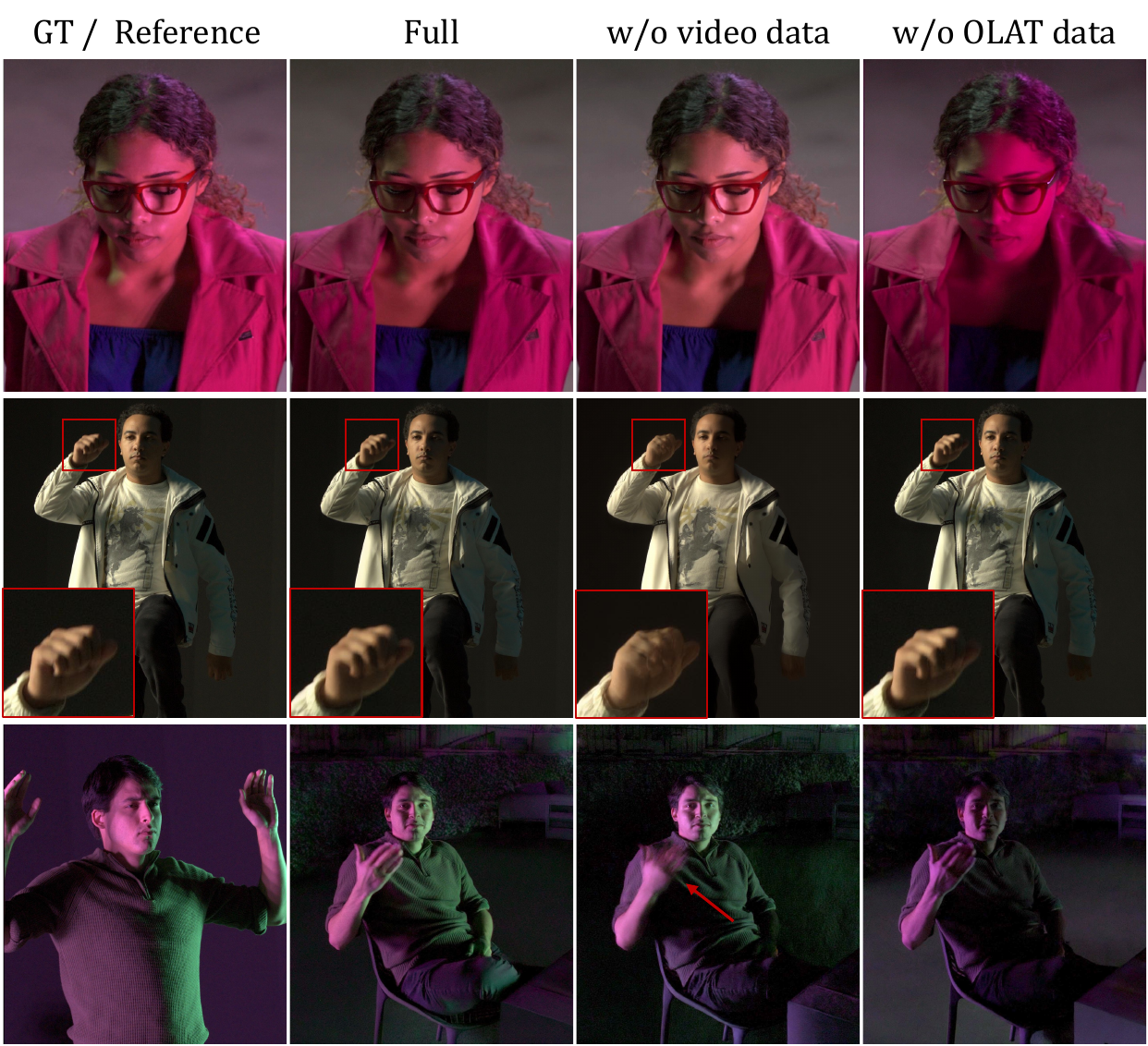}
    \vspace{-5mm}
    \caption{\textbf{Ablation of different types of training data.} Without bi-packed video data, the model fails to produce plausible results when trained only on pseudo motion derived from static OLAT captures. Without OLAT data, the model cannot accurately relight scenes under extreme lighting conditions.}
    \label{fig:ablation_data}
\end{figure}

\paragraph{Ablation on training data}
We evaluate the necessity of using bi-packed video captures and static OLAT captures (w/o video data and w/o OLAT data).
Visual results are shown in Fig.~\ref{fig:ablation_data}. 
Both data types are indeed necessary to achieve photorealistic, robust, and accurate relighting. 
Without bi-packed video data (w/o video data), the model is trained purely on pseudo-videos generated from transformed static images and fails to generalize to realistic human dynamics. 
On the other hand, static OLATs provide a broader distribution of lighting conditions. Without them (w/o OLAT data), the model struggles to produce plausible relighting results, particularly under extreme lighting conditions such as highly directional illumination.

\paragraph{Ablation of lighting conditioning}

\begin{figure}
    \centering
    \includegraphics[width=\linewidth]{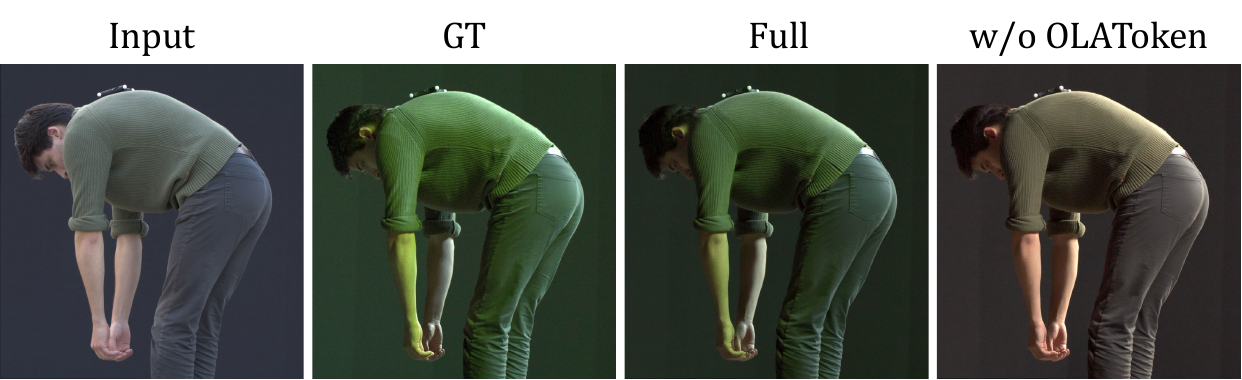}
    \vspace{-6mm}
    \caption{\textbf{OLAToken conditioning.} Our OLAToken conditioning method helps improve the lighting accuracy.}
    \label{fig:abaltion_olatoken}
\end{figure}

\begin{figure}
    \centering
    \includegraphics[width=\linewidth]{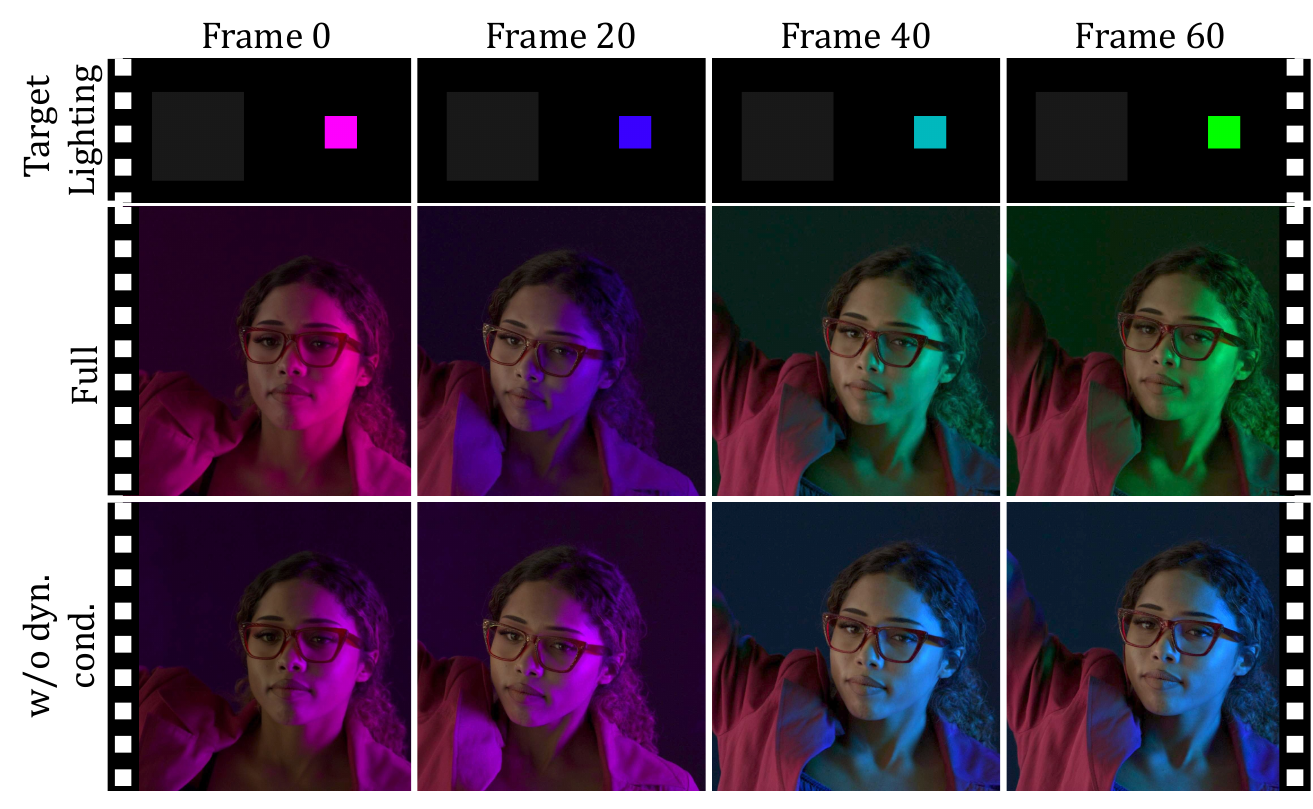}
    \vspace{-6 mm}
    \caption{\textbf{Dynamic lighting conditioning.} Without using dynamic lighting conditioning, lighting conditions tend to leak across frames.}
    \label{fig:ablation_dyncond}
    % \vspace{-2 mm}
\end{figure}

We evaluate the effectiveness of OLAToken conditioning by replacing it with an HDRI-based conditioning scheme (w/o OLAToken). 
Following prior work \cite{iclight,luxpostfacto}, we implement the HDRI-based conditioning by encoding a 2D HDRI into a 1D embedding using a shallow convolutional neural network (CNN), which is then fed to the DiT backbone via cross-attention. 
To support dynamic lighting, we encode HDRI sequences frame by frame and apply dynamic lighting attention. 
As qualitative results show in Fig.~\ref{fig:abaltion_olatoken}, without OLAToken, the relighting results are less accurate than the ground truth, which is also reflected in the quantitative results reported in Tab.~\ref{tab:ablation}.

We further evaluate our dynamic lighting conditioning by replacing it with a straightforward temporal conditioning scheme (w/o dyn. cond.).
Specifically, we remove the temporal attention mask, and add RoPE \cite{rope} time embedding to each lighting token. 
%In this way, each frame is softly constrained by its corresponding lighting. 
We show an example of dynamic lighting with smoothly changing light colors in Fig.~\ref{fig:ablation_dyncond}.
The results show that without dynamic lighting attention, lighting conditions at different frames leak into each others.
\li{This indicates that the model fails to correctly reason about the correspondence between the video frames and the lighting sequence relying solely on time embedding.}
% This indicates that the model fails to reason on the correct conditioning correspondence solely on the time embedding.
% \MINGMING{what does conditioning correspondence mean?}

\begin{figure}
    \centering
    \includegraphics[width=\linewidth]{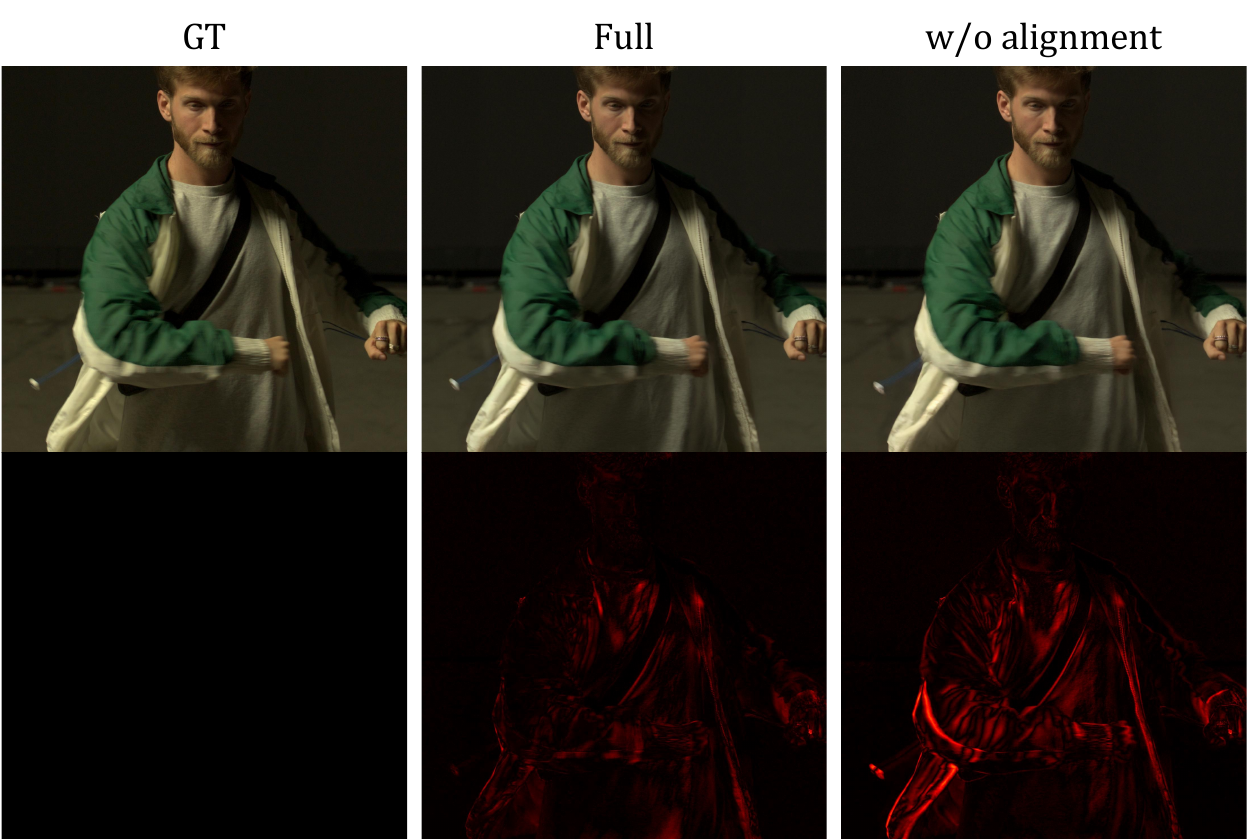}
    \vspace{-5mm}
    \caption{\textbf{Effectiveness of alignment.} Without alignment of video relighting pairs, the model achieves similar quality but suffers from misalignment from the input.}
    \label{fig:ablation_align}
\end{figure}

\camready{
\paragraph{Ablation of frame alignment}
In bi-pack video capture, frames under different lighting conditions are recorded with a slight temporal offset. 
We evaluate the importance of aligning input and target frames when generating the training data for video relighting by removing this step (w/o alignment). 
The results are shown in Tab.~\ref{tab:ablation} and Fig.~\ref{fig:ablation_align}. 
Surprisingly, even without alignment, the model produces visually plausible relighting results. However, the relit outputs exhibit temporal misalignment with respect to the input video.
In practical production settings, such misalignment can complicate downstream compositing, making frame alignment a necessary component of our pipeline.
}

\paragraph{Ablation of pretraining prior}

\begin{figure}
    \centering
    \includegraphics[width=\linewidth]{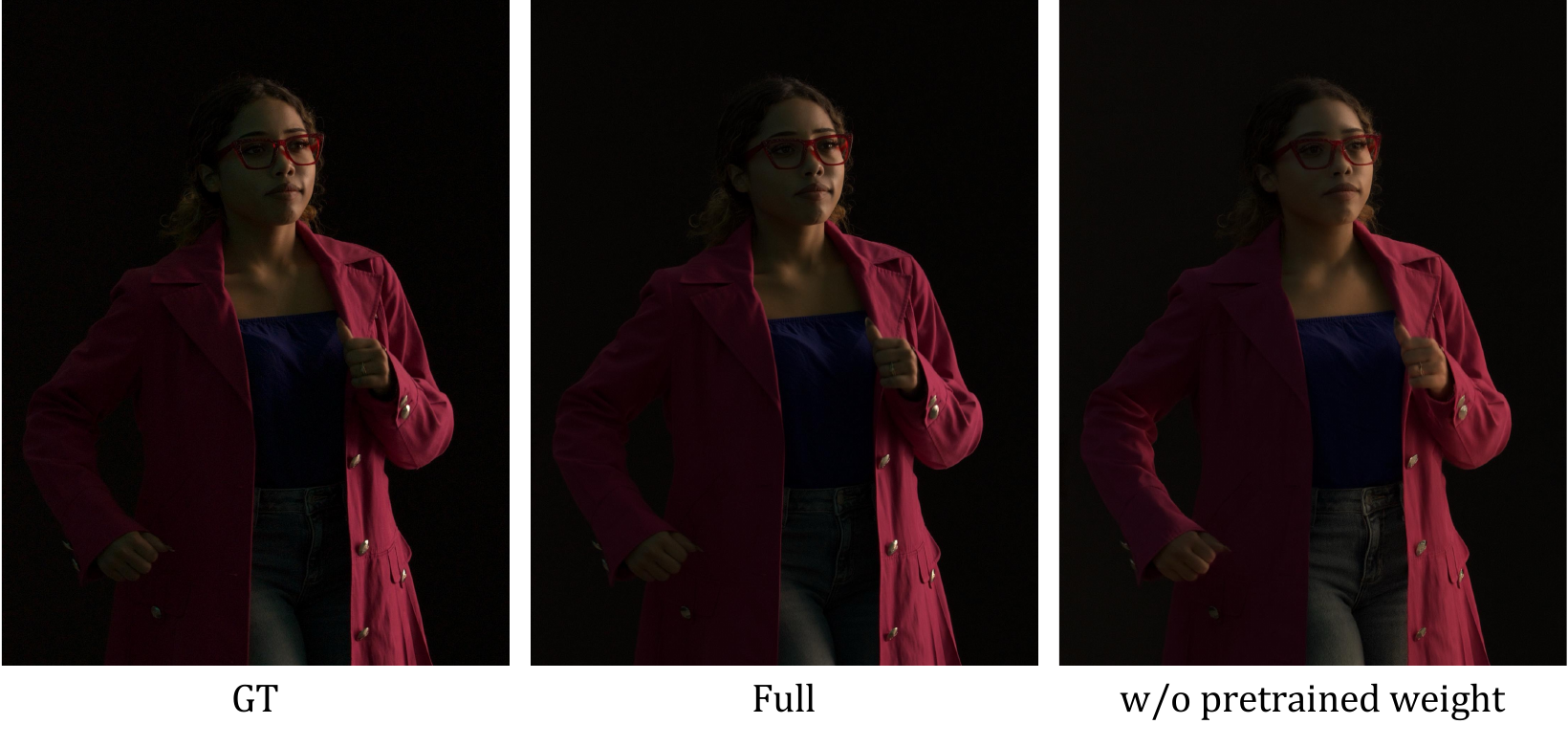}
    \vspace{-5mm}
    \caption{\textbf{Effectiveness of pretrained weight.} Without loading the pretrained weight, the relighting result tends to be blurry with fewer specular reflections and lower relighting accuracy.}
    \label{suppfig:pretrain}
\end{figure}

We evaluate the effectiveness of using pretrained DiT weights by comparing our approach against a model trained from random initialization. The numerical results are shown in Tab.~\ref{tab:ablation}, and one example is shown in Fig.~\ref{suppfig:pretrain}. 
Without prior knowledge embedded in the pretrained weights, the model produces less accurate relighting results characterized by blurry details and diffused specular highlights.

\camready{
\paragraph{Ablation of other WAN variants}
We further evaluate our method using a different diffusion backbone, WAN2.1 1.3B, as shown in Tab.~\ref{tab:ablation}. This variant achieves comparable performance, yielding slightly higher PSNR and SSIM, and lower LPIPS. This suggests that our method is largely backbone-agnostic. However, WAN2.1 uses a VAE with a smaller compression ratio, making it approximately $5\times$ slower inference at the same resolution despite having fewer parameters. We do not evaluate the 14B variant due to its prohibitively expensive train cost and substantial memory requirements. Overall, WAN2.2 5B provides a favorable trade-off between quality, training cost and inference efficiency.
}

\subsection{Additional Analysis}

\begin{figure}
    \centering
    \includegraphics[width=\linewidth]{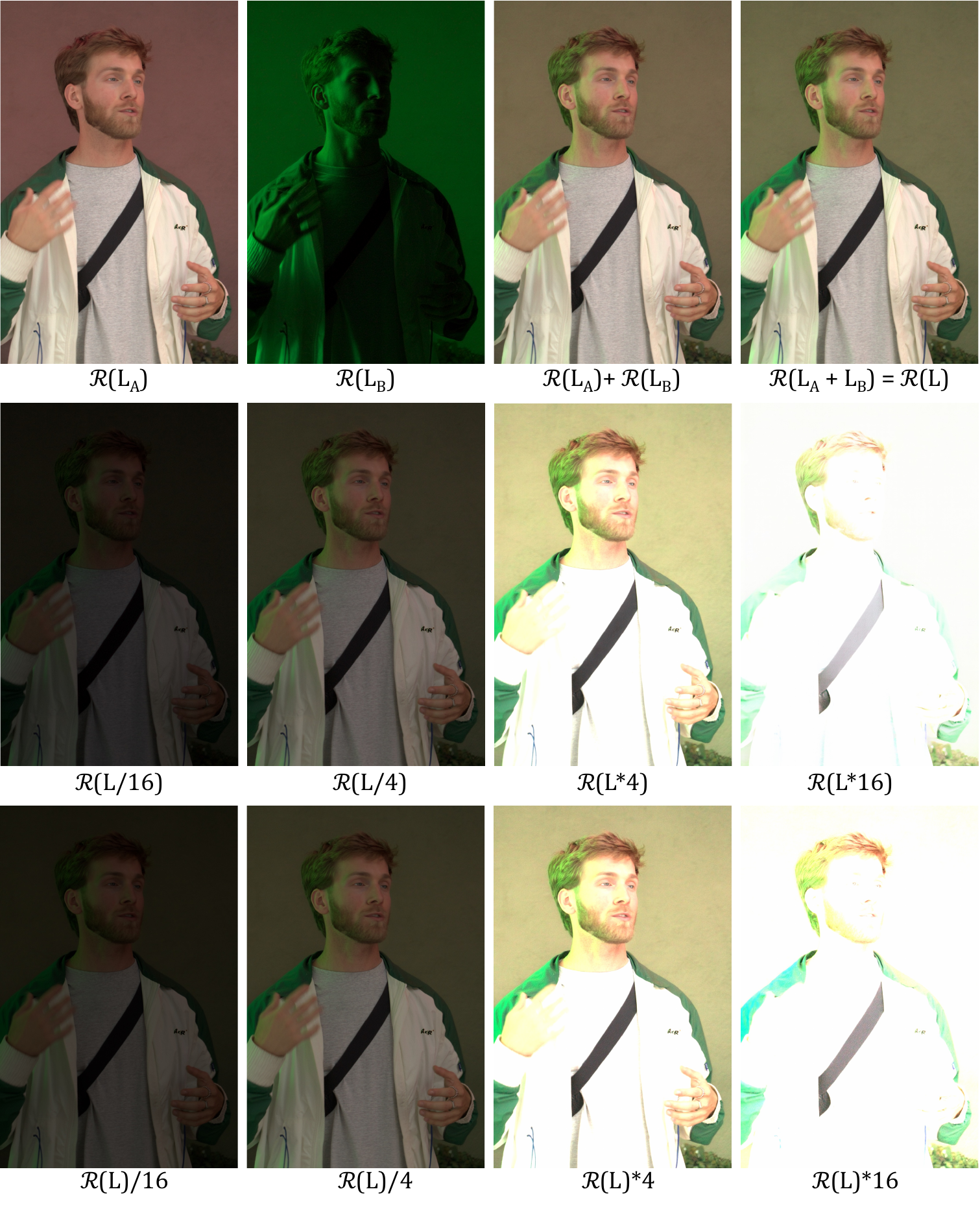}
    \vspace{-6mm}
    \caption{\textbf{Linearity test of our relighting model.} Our relighting model produces visually linear results with respect to the input lighting conditions. Here $
    \mathcal{R}$ indicates our video relighting model.}
    \label{suppfig:linearity}
\end{figure}

\camready{
\paragraph{Linearity of relighting}
Since the lighting transport is linear, we can evaluate the accuracy of lighting control by testing the linearity of the output image $\mathcal{R}(L)$ with respect to the input lighting condition $L$.
For brevity, we denote the relighting model as $\mathcal{R}(L)$ and omit the input image.
We assess linearity through two tasks, light combination and exposure scaling.
If the relighting function $\mathcal{R}$ is linear with respect to $L$, it should satisfy $\mathcal{R}(L_A + L_B) = \mathcal{R}(L_A) + \mathcal{R}(L_B)$, and $\mathcal{R}(\alpha L) = \alpha \mathcal{R}(L)$, where $\alpha$ is a scalar.
Accordingly, we evaluate the linearity by comparing $\mathcal{R}(L_A + L_B)$ with $\mathcal{R}(L_A) + \mathcal{R}(L_B)$ and $\mathcal{R}(\alpha L) $ with $\alpha \mathcal{R}(L)$.
The results are shown in Fig.~\ref{suppfig:linearity}.
As the model is trained in the sRGB color space, all outputs are converted to linear space for evaluation and then back to sRGB for visualization.
The results indicate that our method exhibits near-linear behavior under both lighting addition and exposure scaling.
}

\begin{table}[t]
\centering
\small
\caption{\camready{\textbf{Repeatability evaluation.} Performance remains consistent across models trained on different subsets, demonstrating strong repeatability. Training on fewer captures results in only a minor degradation in quality.}}
\label{tab:repeatibility} 
\vspace{-3mm}
\begin{tabular}{lccc}
\toprule
Training data & PSNR$\uparrow$ & SSIM$\uparrow$ & LPIPS$\downarrow$ \\
\midrule
All captures & \textbf{21.91} & \textbf{0.9462} & \textbf{0.05070} \\
Subset A & 21.47 & 0.9436 & 0.05418 \\
Subset B & 21.52 & 0.9431 & 0.05499 \\

\bottomrule
\vspace{-4mm}
\end{tabular}
\end{table}

\begin{figure}
    \centering
    \includegraphics[width=\linewidth]{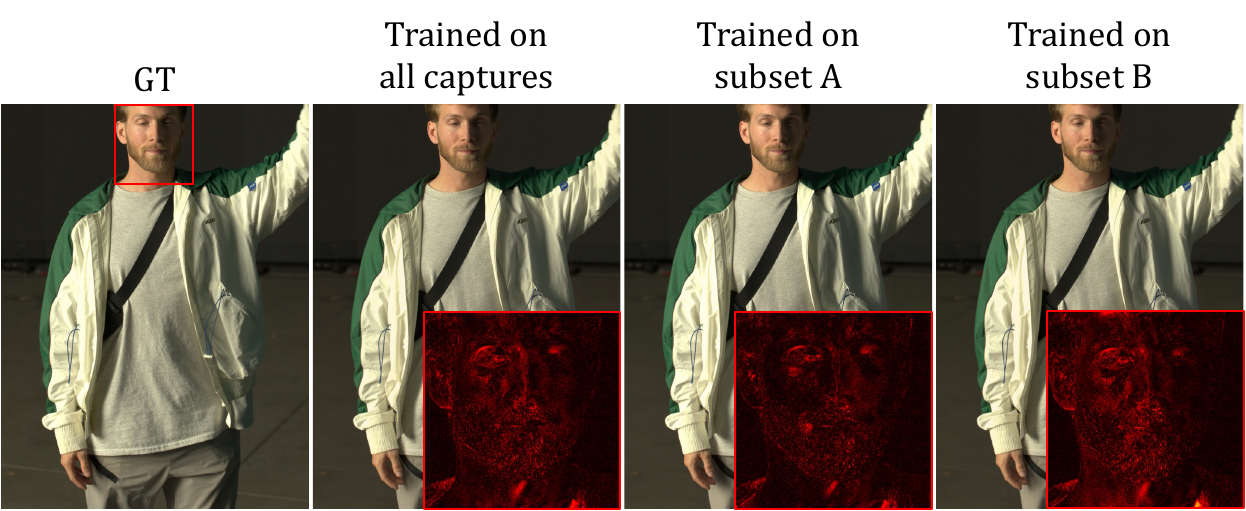}
    \vspace{-4mm}
    \caption{\camready{\textbf{Repeatability across capture subsets.} Models trained on all captures or disjoint subsets produce consistent relighting results. Error maps (insets) show strong pixel alignment of the outputs.}}
    \label{fig:repeatibility}
\end{figure}

\camready{
\paragraph{Repeatability of results}
We assess the repeatability of our method by conducting three independent training experiments. Specifically, we randomly pick two mutually exclusive subsets from all of the captures of a subject, A and B, each containing 6 static OLAT poses and 4 dynamic bi-pack sequences. 
We train separate models on each subset, as well as an additional model using the full set of captures.

Quantitative results are reported in Tab.~\ref{tab:repeatibility}, and qualitative comparisons are shown in Fig.~\ref{fig:repeatibility}. 
The results demonstrate consistent performance across different capture subsets, indicating strong repeatability. 
Using fewer captures leads to a slight degradation in quality, yet the relighting capability remains stable.

In Fig.~\ref{fig:repeatibility}, we additionally visualize error maps. 
The outputs are almost perfectly pixel-aligned with the input.
Minor misalignment (typically on the order of 1-3 pixels around some edges) can occasionally occur, likely due to the spatial compression of the VAE.
}

\begin{figure}
    \centering
    \includegraphics[width=\linewidth]{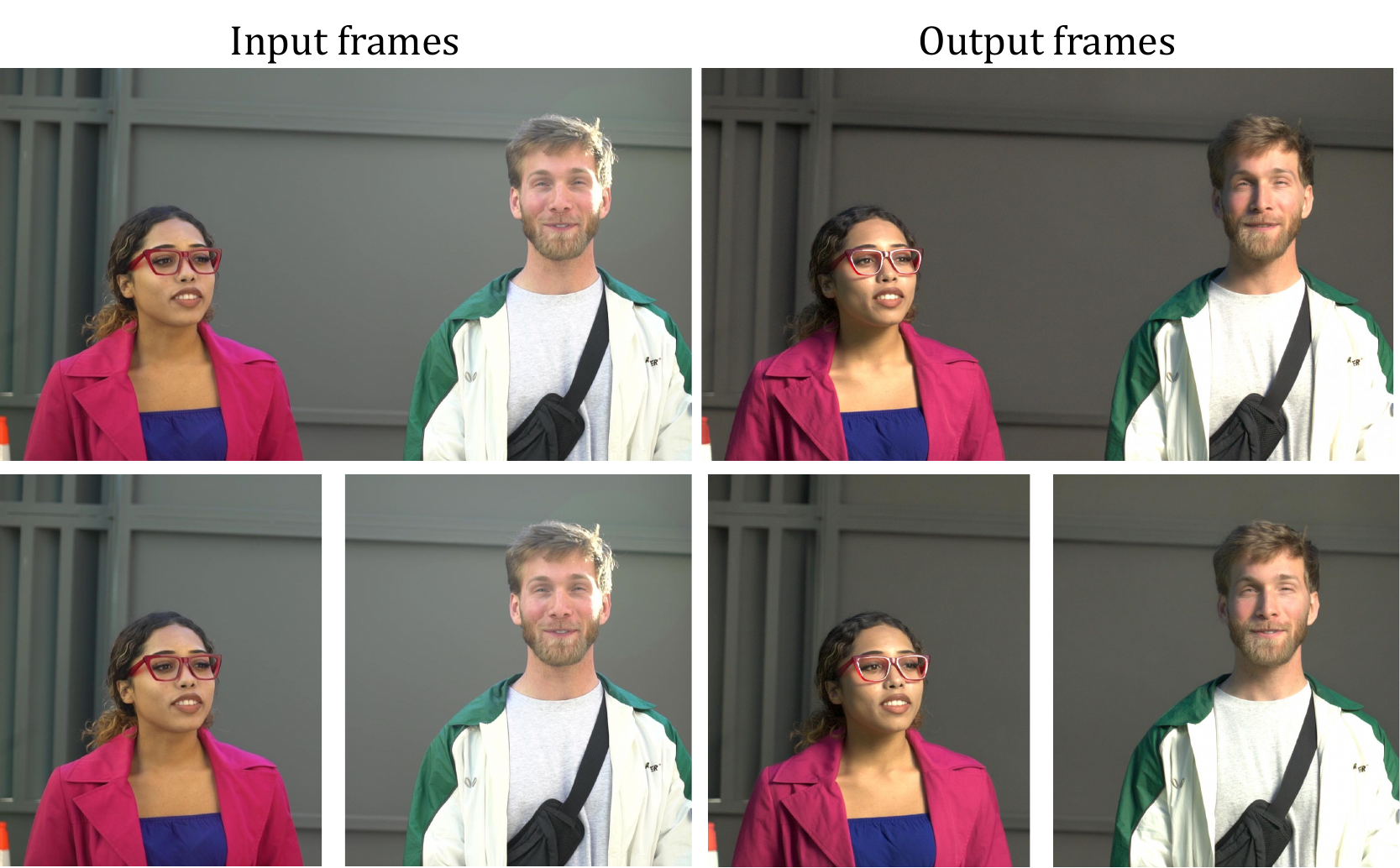}
    \caption{\textbf{Relighting results on multi-subject shot.} We compare relighting results on a two-subject video (first row), with relighting them individually by cropping (second row). The model is able to generalize to multi-subject shot even if it's only trained on single-subject data.}
    \label{suppfig:twoactor}
\end{figure}

\camready{
\paragraph{Multiple subjects inference}
Although our training data consists exclusively of single-subject sequences, we test our relighting model on a two-subject scene, as shown in Fig.~\ref{suppfig:twoactor}. 
We compare against two individual inference results on crops of each subject, and find the results to be visually consistent.
This suggests that our model generalizes naturally to multi-subject scenarios.
However, more complex interactions—such as severe occlusions and inter-subject shadow casting—remain challenging. We leave these cases for future work.
}

% %%%%%%%%%%%%%%%%%%%%%%%%%%%%%%%%%%%%%%%%%%%%%%%%%%%%

\section{Limitations, Future Work and Conclusion}

\begin{figure}
    \centering
    \includegraphics[width=\linewidth]{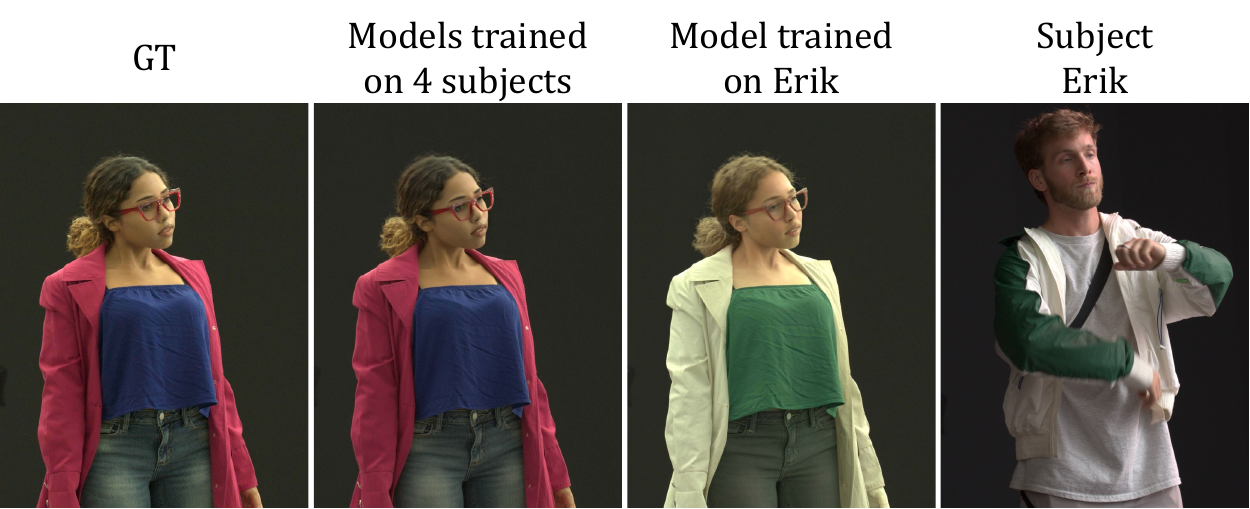}
    \vspace{-6mm}
    \caption{\textbf{Relighting results from out-of-distribution identities.} When inferred on unseen identities, the model tends to bake in subject-specific features such as cloth colors, hair color, and facial features.}
    \label{fig:subjectspecific}
\end{figure}

We presented a new framework for photo-real, temporally consistent human performance relighting. 
We propose a new video capture process to train a video diffusion-based relighting model with a novel lighting conditioning mechanism achieving accurate relighting. 
We achieve the best relighting quality compared to baselines, and our ablation studies show the effectiveness of our design choices.

Our method still has a few limitations.
First, our \LightDome only lights from the upper hemisphere and lights are a few meters from the subjects.
Therefore, our model is not able to relight with light from the lower hemisphere, or with near-field lighting \cite{Jones:SSV:2006}.
Second, our method is subject-specific and not designed to relight unseen identities.
We show one example in Fig.~\ref{fig:subjectspecific}, where we train our model for one subject, and infer on a different subject.
Interestingly, it still produces reasonable relighting, but it transfers features from the training subject to the new subject, including the color of the jacket and hair, and facial features like the beard.
This indicates that the model may learn to relight based on semantic features from the input videos, rather than purely overfitting to relight one subject.

For future work, we hope to scale the data capture to more identities, performances, props, and scenes to train a generalized subject and scene relighting model.

\begin{acks}
We would like to thank Jeffrey Shapiro for his ongoing support; Alek Nieberlein and Samuel Price for stage operation; Jennifer Lao for performer coordination; Lauren Wilson for stage scheduling; Emmett Steven for infrastructure support; Scott McDonald for Red camera control; the Software Department for stage software development, and our performers: Pablo Salamanca, David George, Erik Patten and Anica Rose.
\end{acks}

\clearpage

\bibliographystyle{ACM-Reference-Format}
\bibliography{bibliography}

\end{document}